%% file: main.tex
\documentclass[letterpaper, 10 pt, conference]{ieeeconf}  % Comment this line out if you need a4paper

\IEEEoverridecommandlockouts                              % This command is only needed if 

\usepackage{graphics} % for pdf, bitmapped graphics files
\usepackage{epsfig} % for postscript graphics files
\usepackage{mathptmx} % assumes new font selection scheme installed
\usepackage{times} % assumes new font selection scheme installed
\usepackage{amsmath} % assumes amsmath package installed
\usepackage{amssymb}  % assumes amsmath package installed
\usepackage{multirow}
\usepackage{booktabs}
\usepackage{overpic}
\usepackage{makecell}
\usepackage{threeparttable}
\usepackage{url}
\usepackage[bookmarks=true]{hyperref}

\title{\LARGE \bf
MAIA: Anticipating Interaction Anomalies in Contact-Rich Manipulation via Multimodal World Models
}

\author{Yi Qiao$^{1}$, Maolin Zheng$^{1}$, Haoran Sun$^{1}$, Lujia Dong$^{1}$, and Wenzhao Lian$^{1,*}$%
\thanks{$^{1}$School of Artificial Intelligence, Shanghai Jiao Tong University}%
\thanks{*Corresponding author.}%
}

\begin{document}

% \IEEEaftertitletext{%
%     \centering
%     \includegraphics[width=0.7\textwidth]{images/image.png}\\[-1mm]
%     {\footnotesize
%     Fig.~1. Overview of our framework. 
%     Our method ... 
%     }\\[2mm]
% }

\makeatletter
\let\@oldmaketitle\@maketitle% Store \@maketitle
\renewcommand{\@maketitle}{%
    \@oldmaketitle%
    % \vspace{-10pt}%   % for no author
    {\centering
        \includegraphics[width=\textwidth]{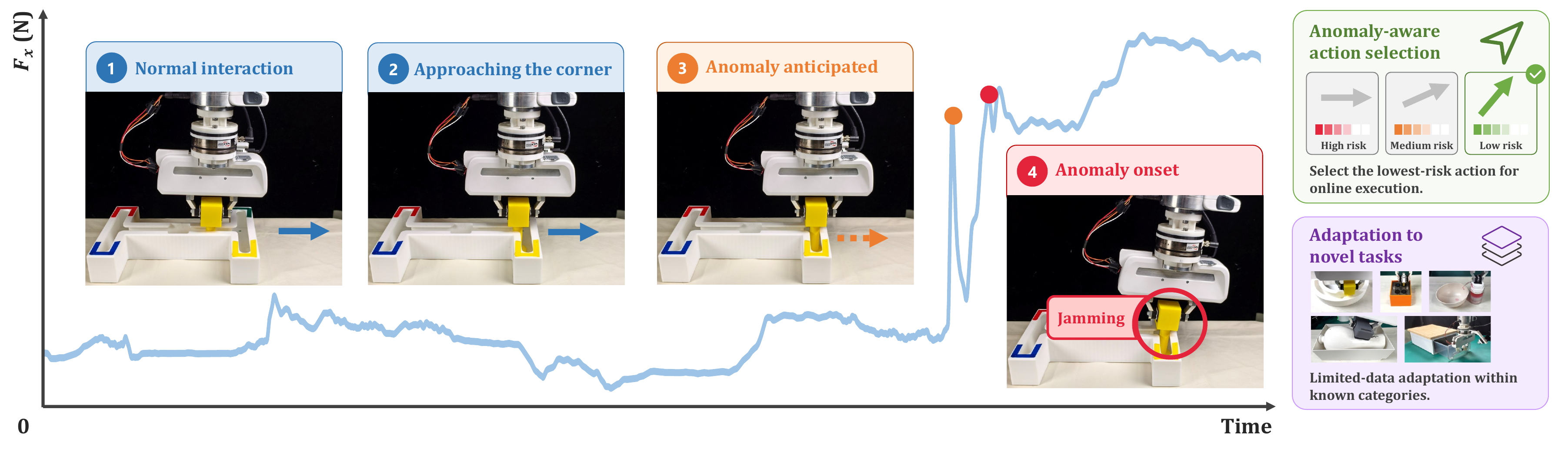}
        \par
    }    
    \refstepcounter{figure} 
    \footnotesize{Fig.~\thefigure. ~\textbf{Anticipating anomalous interactions in contact-rich manipulation.} A representative path-constrained manipulation sequence progresses from nominal interaction (1–2), through an early warning (3), to jamming onset (4). The curve shows the measured force $F_x$, with orange and red markers indicating the anticipation time and anomaly onset, respectively. By integrating visual observations, planned actions, and force feedback, our framework anticipates interaction anomalies, supports anomaly-aware action selection, and adapts to novel tasks with limited data.}%
    \label{fig:teaser} 
    \vspace{-3pt}%
    }% ... an image
\makeatother

\maketitle
\thispagestyle{empty}
\pagestyle{empty}

%%%%%%%%%%%%%%%%%%%%%%%%%%%%%%%%%%%%%%%%%%%%%%%%%%%%%%%%%%%%%%%%%%%%%%%%%%%%%%%%
\begin{abstract}
% Reliable contact-rich manipulation requires timely responses to execution deviations before they escalate and compromise task completion. However, reactive monitoring identifies such deviations only after they become evident, narrowing the opportunity for corrective action. Anticipating these deviations is challenging because physical interaction is only partially reflected in visual observations, while planned actions specify intended motion rather than actual physical consequences. 
% To address this, we propose MAIA, a multimodal framework for predictive anomaly detection. The Vision–Action–Force World Model (VAF-WM) learns nominal interaction dynamics and forecasts future visual and force responses, while a lightweight anomaly head uses these representations to assess whether the planned actions will lead to anomalous interactions. 
% We further introduce DATASET, a real-world multimodal dataset comprising 2,852 trajectories across 18 contact-rich manipulation tasks with fine-grained temporal anomaly annotations.
% Experiments across both seen and novel tasks show that MAIA outperforms representative baselines in anomaly discrimination while providing earlier warnings. Anomaly-aware action selection further improves overall task success without policy retraining.

Reliable contact-rich manipulation requires detecting execution deviations early enough for corrective action. Yet most monitoring methods are reactive, identifying anomalies only after they become observable. Anticipating such anomalies is challenging because visual observations only partially reveal physical interaction, while planned actions describe intended motion rather than their actual physical consequences. 
We address this problem with MAIA, a multimodal framework for anomaly anticipation. Its vision-action-force world model learns nominal interaction dynamics and predicts the visual and force responses of planned actions, while a lightweight anomaly head predicts impending interaction anomalies. 
To enable systematic evaluation, we introduce CRIA, a real-world multimodal dataset of 2,852 trajectories spanning 18 contact-rich manipulation tasks, with frame-level annotations of anomaly onset and evolution.
Experiments on both seen and novel tasks show that MAIA achieves more accurate anomaly discrimination than representative baselines while providing earlier warnings, with bACC scores of 0.765 and 0.710 and median anticipation lead times of 28.0 and 25.1 steps, respectively. Moreover, using these predictions for anomaly-aware action selection improves end-to-end task success from 52.6\% to 61.5\% without retraining the underlying manipulation policy.
These results highlight the value of modeling future physical consequences for proactive anomaly anticipation.
View our project at \url{https://joringell.github.io/project/MAIA/}
\end{abstract}

%%%%%%%%%%%%%%%%%%%%%%%%%%%%%%%%%%%%%%%%%%%%%%%%%%%%%%%%%%%%%%%%%%%%%%%%%%%%%%%%

\input{sections/1_introduction}

\input{sections/2_related_work}
\input{sections/3_problem_formulation}
\input{sections/4_dataset}
\input{sections/5_method}
\input{sections/6_experiments}

\input{sections/7_conclusion}

%%%%%%%%%%%%%%%%%%%%%%%%%%%%%%%%%%%%%%%%%%%%%%%%%%%%%%%%%%%%%%%%%%%%%%%%%%%%%%%%

%                                   % the textheight of the last page by a suitable amount.
%                                   % This command does not take effect until the next page
%                                   % so it should come on the page before the last. Make
%                                   % sure that you do not shorten the textheight too much.

\bibliographystyle{IEEEtran}
\bibliography{references}

\input{sections/X_appendix}

\end{document}

%% file: sections/1_introduction.tex
\section{Introduction}
% Modern robot manipulation policies can plan and execute increasingly complex behaviors, yet reliable execution in the physical world remains far from guaranteed~\cite{pi05,dp, cosmospolicy}.
Modern robot manipulation policies can plan and execute increasingly complex behaviors, yet even strong policies cannot eliminate execution uncertainty during physical interaction~\cite{pi05,dp, cosmospolicy}.
While policy-generated actions appear reasonable, perception errors, imperfect alignment, object constraints, and environmental variations can cause the resulting interaction to deviate from the intended trajectory.
This uncertainty is especially pronounced in contact-rich manipulation, where similar actions can lead to substantially different physical outcomes depending on how the interaction unfolds~\cite{pang2023global,cheng2022contact}.
Reliable execution therefore requires monitoring the evolving robot–environment interaction and responding to deviations before they escalate and compromise task progress.

Existing approaches to robot execution monitoring predominantly focus on reactive detection of failures and anomalies~\cite{faildetect,SAFE}. By identifying deviations only after observable evidence emerges, they leave little window for intervention. This is particularly limiting for anomalous interactions, which may emerge locally and evolve without producing clear task-level failure signals. 
Modern manipulation policies, however, often generate action sequences over a future horizon, enabling evaluation of their potential consequences in advance~\cite{pi05,dp,act}. 
This motivates anomaly anticipation, which assesses whether planned actions are likely to induce anomalous interactions before they occur, thereby providing an early warning for intervention or replanning, as illustrated in Fig.~\ref{fig:teaser}.

A central challenge in predicting anomalous interactions stems from the partial observability of underlying physical dynamics.
While visual observations provide valuable geometric context and task progress, they offer only indirect evidence of contact forces, friction effects, resistance, and mechanical constraints. 
% Planned actions, meanwhile, represent only the robot's intended motion rather than the actual environmental response.
Planned actions specify the robot’s intended motion, yet do not reveal how the environment will physically respond.
% Force-torque feedback bridges this gap by grounding visual context and action intent in measured physical responses.
% Together, vision and action alone cannot fully determine the future interaction state. 
Thus, similar visual states and actions can lead to substantially different interaction outcomes.
Force–torque feedback reduces this ambiguity by providing direct measurements of physical responses.
% Anticipating anomalous interactions therefore requires modeling future multimodal physical responses under planned actions.
This calls for modeling future multimodal physical responses under planned actions to anticipate anomalous interactions.

Driven by these insights, we formulate contact-rich anomaly anticipation as prospective action evaluation: predicting whether planned actions will cause the robot–environment interaction that deviates from its expected dynamics. 
% We instantiate this formulation in MAIA, a multimodal framework that couples interaction forecasting with anomaly estimation. 
We instantiate this formulation in \textbf{MAIA}, a framework for \textbf{M}ultimodal \textbf{A}nticipation of \textbf{I}nteraction \textbf{A}nomalies.
At its core, the vision–action–force world model learns expected interaction dynamics from nominal executions and predicts future visual and force responses under planned actions. A lightweight anomaly head then integrates the resulting predictive representations to estimate whether an anomaly will occur and its temporal prevalence over the future horizon. 

However, evaluating anomaly anticipation requires temporally localized supervision indicating when an anomaly begins and how it evolves, which is not provided by conventional failure-oriented datasets. Thus, we introduce \textbf{CRIA}, a real-world multimodal dataset for \textbf{C}ontact-\textbf{R}ich \textbf{I}nteraction \textbf{A}nomaly anticipation. CRIA comprises 2,852 trajectories spanning 18 contact-rich manipulation tasks, covering nominal and anomalous executions. It provides temporally aligned visual, action, and force/torque signals with frame-level annotations of anomaly onset and evolution.

Built on this data, MAIA learns interaction dynamics rather than task-specific failure templates. This interaction-centric representation captures shared dynamics across diverse contact regimes while remaining sensitive to task-specific deviations, supporting transfer to novel tasks with limited adaptation. 
Additionally, because model predictions are conditioned on planned actions, MAIA can evaluate candidate action sequences before execution, extending anomaly anticipation from passive monitoring to proactive action selection.
% Beyond passive monitoring, its anomaly predictions can guide online action selection.
% , enabling proactive safeguarding.

Our key contributions are summarized as follows:
\begin{itemize}
    % \item We formulate contact-rich anomaly anticipation as prospective action evaluation: predicting whether planned actions will drive the robot–environment interaction away from its expected physical evolution. We instantiate this in MAIA, a Vision–Action–Force framework for anomaly anticipation.
    \item We formulate contact-rich anomaly anticipation as prospective action evaluation and introduce MAIA, a vision–action–force framework for anticipating anomalous interactions.
     
    \item We introduce CRIA, a real-world aligned visuo-action-force dataset for contact-rich manipulation, comprising 2,852 trajectories across 18 tasks with fine-grained temporal anomaly annotations.
    
    \item We demonstrate that MAIA learns transferable predictive interaction representations across diverse contact regimes, achieving bACC of 0.765 on seen tasks and 0.710 on novel tasks. MAIA further enables anomaly-aware action selection, improving real-robot success from 52.6\% to 61.5\% without policy retraining.
\end{itemize}

%% file: sections/2_related_work.tex
\section{Related Work}
\subsection{Robot Failure and Anomaly Detection}
% Robot failure and anomaly detection has evolved from identifying deviations in observed execution toward more general runtime monitoring and predictive assessment. 
% Robot failure and anomaly detection is evolving from reactive recognition toward proactive anticipation.

% Reliable robot manipulation requires timely detection of execution deviations that may indicate anomalous interactions or impending task failures.
Robot execution monitoring aims to detect deviations that may indicate anomalous interactions or impending task failures.
A common line of work formulates failure monitoring as out-of-distribution (OOD) detection, learning representations of nominal execution and detecting departures from the learned distribution~\cite{faildetect,rnd,rcnf}.
With the emergence of vision-language-action policies, recent methods exploit temporal action consistency~\cite{Hide-and-Seek,ActFovea}, internal policy representations~\cite{SAFE,VLA-FAIL}, and semantic reasoning~\cite{guardian} for runtime monitoring and failure characterization.
More recently, predictive methods have begun to assess the consequences of planned actions through learned future representations or world-model rollouts, offering earlier evidence of potential failures~\cite{foresight,foundationalwm,forewarn,surgicalwm}.
However, these approaches seldom explicitly model physical interaction dynamics under future actions. In contrast, we anticipate anomalies by modeling how the interaction is expected to evolve from visual observations, planned action sequences, and force-torque feedback.

\subsection{World Models for Robot Interaction}
% World models learn predictive representations of robot-environment interactions and have been widely used for robot planning and control, ranging from pixel-space visual prediction to latent-space dynamics modeling~\cite{visualforesight,dinowm,vjepa2}.
% Recent work further leverages imagined futures to support robot action generation, evaluation, and selection~\cite{cosmospolicy,ctrlworld,rise,world4rl,factwam,forewarn}.
% For contact-rich manipulation, multimodal world models have begun to incorporate tactile or force sensing to capture physical responses that are difficult to infer from visual observations alone~\cite{vitacworld,tacforesight,dreamtac,OmniVTA,contactworld,fawam}.
% While these multimodal world models have primarily been applied to policy learning and action refinement, their use for anomaly monitoring remains relatively underexplored. We therefore leverage their predictive capability to develop a vision-action-force world model for anticipating deviations from expected interaction dynamics under planned actions.
World models learn predictive representations of robot-environment interactions to support planning and control, modeling future observations in pixel space or latent dynamics space~\cite{visualforesight,dinowm,vjepa2}. 
Building on these predictive capabilities, recent methods leverage imagined futures to guide action generation, evaluation, and selection~\cite{cosmospolicy,forewarn,rise,factwam}. 
To better capture contact-rich physics, multimodal world models have increasingly incorporated force or tactile sensing alongside visual inputs~\cite{vitacworld,tacforesight,OmniVTA,TacWAM,fawam}.
However, existing approaches predominantly focus on predicting future visual scenes or generating next actions, rather than explicitly modeling the physical consequences resulting from planned actions. We instead forecast future visual and force responses conditioned on these actions, revealing how the physical interaction is expected to evolve and enabling anomaly anticipation before deviations become 
fully observable.
% providing a representation of the resulting physical interaction that enables anomaly anticipation before deviations become fully observable.

%% file: sections/3_problem_formulation.tex
\section{Problem Formulation}
Robot manipulation naturally unfolds through continuous interaction with the environment, where visual observations, robot actions, and physical responses jointly characterize the evolving interaction state.
Under normal execution, the interaction state evolves toward the task goal in accordance with the underlying task semantics and physical constraints. 
We refer to such task-oriented evolution as the \textbf{expected interaction dynamics}, and define a manipulation anomaly as an observable deviation from these dynamics during execution. 
% Notably, an anomaly is not equivalent to episode-level task failure: an anomalous interaction may be recovered from, while a failed episode may remain normal until the actual anomalous event occurs.
% Notably, an anomaly is not equivalent to an episode-level task failure: an anomalous interaction may be transient or recoverable, whereas a failed trajectory remains nominal until the onset of the anomalous event.
Notably, an anomaly is distinct from episode-level task failure: the former describes a local deviation in interaction dynamics, whereas the latter reflects the eventual task outcome.

%  for with author
\begin{figure*}
    \centering
    \vspace{5pt}
    \includegraphics[width=\linewidth]{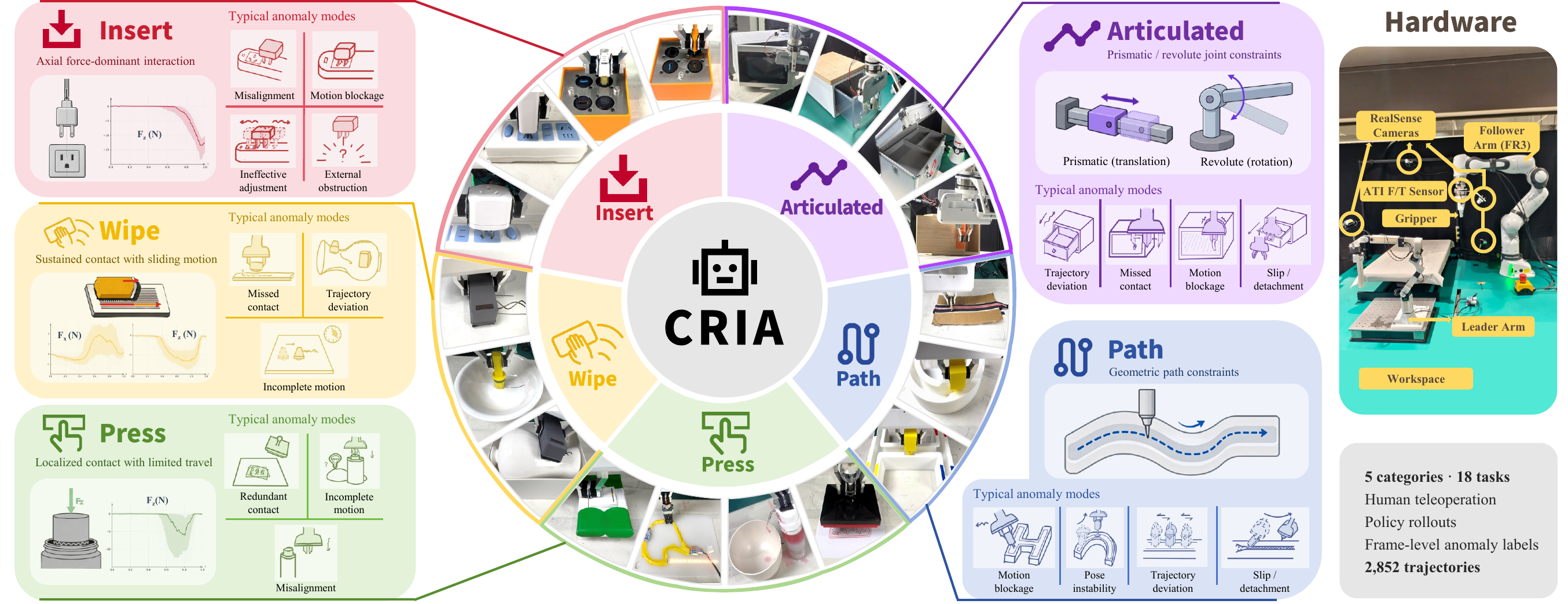}
    \caption{\textbf{CRIA overview.} The central ring presents all 18 tasks grouped by interaction category, while the surrounding panels illustrate category-specific interaction characteristics and typical anomaly modes. The hardware setup used for data collection is shown on the right.}
    \label{fig:dataset}
    \vspace{-3.5mm}%
\end{figure*}

Rather than detecting anomalies only after they have manifested, our goal is to predict whether the robot will enter an anomalous interaction state under the future actions proposed by the policy. Let $z_\tau \in \{0,1\}$ indicate whether an anomalous interaction occurs at timestep $\tau$.  At the current timestep $t$, given a history of visual observations and force responses $H_t = \{V_{t-T_v+1:t},\, F_{t-T_f+1:t}\}$, a future action sequence $A_{t+1:t+K}$ together with a language task description $L$, we predict two complementary quantities over the next $K$ steps:
\begin{equation}
y_t = \mathbb{I}\left[\sum_{k=1}^K z_{t+k} > 0\right], \quad r_t = \frac{1}{K}\sum_{k=1}^K z_{t+k},
\end{equation}
where $y_t \in \{0,1\}$ indicates whether any anomaly occurs, while
$r_t \in [0,1]$ measures its temporal prevalence. 

%% file: sections/4_dataset.tex
%  for no author
% \begin{figure*}
%     \centering
%     \vspace{5pt}
%     \includegraphics[width=\linewidth]{images/dataset.pdf}
%     \caption{\textbf{CRIA overview.} The central ring presents all 18 tasks grouped by interaction category, while the surrounding panels illustrate category-specific interaction characteristics and typical anomaly modes. The hardware setup used for data collection is shown on the right.}
%     \label{fig:dataset}
%     \vspace{-3.5mm}%
% \end{figure*}   

\section{Dataset}
Existing robot datasets are not designed to evaluate anomaly anticipation: multimodal manipulation datasets lack temporally localized anomaly supervision, while failure-oriented datasets rarely capture the physical interaction signals needed to predict how anomalies emerge. Table~\ref{tab:dataset_comparison} compares representative datasets along these dimensions.

% To fill this gap, we introduce CRIA, a real-world multimodal dataset for contact-rich interaction anomaly anticipation. CRIA comprises 2,852 trajectories across 18 tasks and 5 task categories, collected from both human teleoperation and autonomous policy rollouts. Each trajectory synchronizes multi-view visual observations, robot states and actions, and high-frequency 6-DoF force/torque sensing. Beyond episode-level outcomes, CRIA provides fine-grained temporal annotations that localize anomalous events throughout execution, preserving their emergence, duration, and potential recovery. 
To fill this gap, we introduce CRIA, a real-world multimodal dataset for contact-rich interaction anomaly anticipation. CRIA comprises 2,852 trajectories across 18 tasks and 5 task categories, of which 1,817 contain at least one annotated anomalous event, with synchronized visual, robot state/action, and force/torque observations. Beyond episode-level outcomes, CRIA provides fine-grained temporal annotations that localize anomalous events throughout execution, preserving their emergence, duration, and potential recovery. 

\begin{table*}[t]
\centering
\vspace{5pt}
\begin{threeparttable}
\caption{Comparison with representative multimodal manipulation and failure/anomaly datasets.}
\label{tab:dataset_comparison}
\begin{tabular}{lcccccc}
\toprule
\textbf{Dataset} & \textbf{Physical Sensing} & \textbf{Anomaly Source} & \textbf{Real Anomalous Traj.} & \textbf{Anomaly Supervision} & \textbf{Granularity} \\
\midrule
RH20T~\cite{rh20t} & \checkmark & Human & $\sim$10k & Outcome & Episode \\
REASSEMBLE~\cite{REASSEMBLE} & \checkmark & Human & 516 & Outcome & Segment \\
Hoi!~\cite{hoi} & \checkmark & -- & -- & -- & -- \\
FreeTacMan~\cite{freetacman} & \checkmark & -- & -- & -- & -- \\
\midrule
SOAR~\cite{soar} & $\times$ & Policy & 20,564 & Outcome & Episode \\
RoboFAC~\cite{robofac} & $\times$ & Human + Injected & 480 & Outcome + Type + Explanation & Subtask \\
RoboFail~\cite{robofail} & $\times$ & Human + Injected & 30 & Type + Explanation & Segment \\
GuardianFail-36k~\cite{guardian} & $\times$ & Policy + Injected & 0 & Outcome + Type & Subtask \\
\midrule
\textbf{CRIA(Ours)} & \checkmark & \textbf{Human + Policy} & \textbf{1817} & \textbf{Outcome + Anomaly State} & \textbf{Frame} \\
\bottomrule
\end{tabular}
\begin{tablenotes}
\footnotesize
\item ``Physical Sensing'' indicates whether force/torque or tactile sensing is available.
% \item ``Real anomalous tracjectories'' are collected from real robot executions without offline anomaly injection.
\end{tablenotes}
\end{threeparttable}
\vspace{-4mm}
\end{table*}

\subsection{Task Taxonomy}
We organize 18 contact-rich manipulation tasks into five categories: insertion, pressing, wiping, path-constrained manipulation, and articulated-object manipulation. The categories encompass distinct manipulation constraints and contact patterns, while tasks within each category share similar interaction dynamics but vary in object geometry and task semantics. Despite these differences, recurring anomaly patterns emerge across categories, suggesting shared failure modes in contact-rich interactions, as illustrated in Fig.~\ref{fig:dataset}.

\subsection{Data Collection}
All trajectories are collected on a Franka Research 3 (FR3), a 7-DoF robotic manipulator equipped with an ATI Axia80-M8 6-axis force/torque (F/T) sensor at the wrist. Visual observations are captured by three Intel RealSense D435i cameras from wrist, front, and top-down perspectives, with the front-view camera repositioned for certain tasks. Human demonstrations are collected via a leader–follower teleoperation framework based on FACTR~\cite{factr}, providing the operator with real-time force feedback.

For each task, we collect at least 150 trajectories: at least 50 successful human demonstrations, 50 human demonstrations containing diverse anomalous interactions, and 50 autonomous policy rollouts generated by $\pi_{0.5}$~\cite{pi05}. 
% Successful demonstrations provide a reference for nominal interaction dynamics with natural variability. Anomalous demonstrations broaden the coverage of task-relevant anomalies, including severe or persistent cases that are rarely observed during autonomous execution. Policy rollouts complement the human data with natural execution variability, including deviations that arise under autonomous execution.
Successful demonstrations provide nominal interaction dynamics with natural variability. Anomalous demonstrations extend coverage to severe or persistent cases that are unlikely to arise under autonomous policy execution. Policy rollouts complement human data by capturing the policy’s own interaction distribution during autonomous runs.

\subsection{Multimodal Recording and Annotation}
Each trajectory contains visual observations, robot states and actions, and F/T measurements, all timestamped during recording and temporally aligned during preprocessing. 
% Three RGB streams from the wrist, front, and top-down views 
Three RGB streams (wrist, front, top-down)
are recorded at 30 Hz with a resolution of $256\times192$, together with robot joint states, end-effector poses, gripper states, and control actions at the same frequency. F/T measurements are sampled at 120 Hz to preserve high-frequency contact responses.

% We establish a fine-grained temporal annotation protocol that assigns a binary nominal/anomalous label to each video frame. Annotators inspect synchronized visual observations and physical signals to identify the temporal onset and offset of anomalous segments. Annotation boundaries are determined by deviations from expected interaction dynamics, considering both task progress and contact responses.
% Importantly, anomaly labels are assigned according to the interaction process rather than the final task outcome: a successful trajectory may contain transient anomalous interactions followed by recovery, while a failed trajectory may remain nominal before an anomaly emerges.
% To ensure annotation consistency, labeled trajectories are cross-checked by multiple annotators. 
% with ambiguous temporal boundaries resolved through consensus.

We establish a fine-grained temporal annotation protocol that assigns a binary nominal/anomalous label to each video frame. Annotators use synchronized visual observations and physical signals to identify anomalous segments and their frame-level boundaries based on deviations from expected interaction dynamics, considering both task progress and contact responses. Importantly, labels reflect the interaction process rather than the final task outcome: successful trajectories may contain transient anomalies followed by recovery, while failed trajectories may remain nominal before an anomaly emerges. All labels are cross-checked by multiple annotators for consistency.

We additionally annotate the onset of task-relevant interaction to separate manipulation from the preceding approaching phase.
Because this boundary depends on task mechanics, the onset is defined using category-specific criteria rather than a 
uniform contact threshold:
\begin{itemize}
    \item For insertion, pressing, and wiping tasks, the onset is marked when the end-effector reaches the target vicinity to initiate the manipulation, before contact.
    \item For articulated-object and path-constrained tasks, the onset is marked once the task-relevant grasp or contact has been established.
\end{itemize}

%% file: sections/5_method.tex
\begin{figure*}[t]
    \centering
    \vspace{5pt}
    \includegraphics[width=\linewidth]{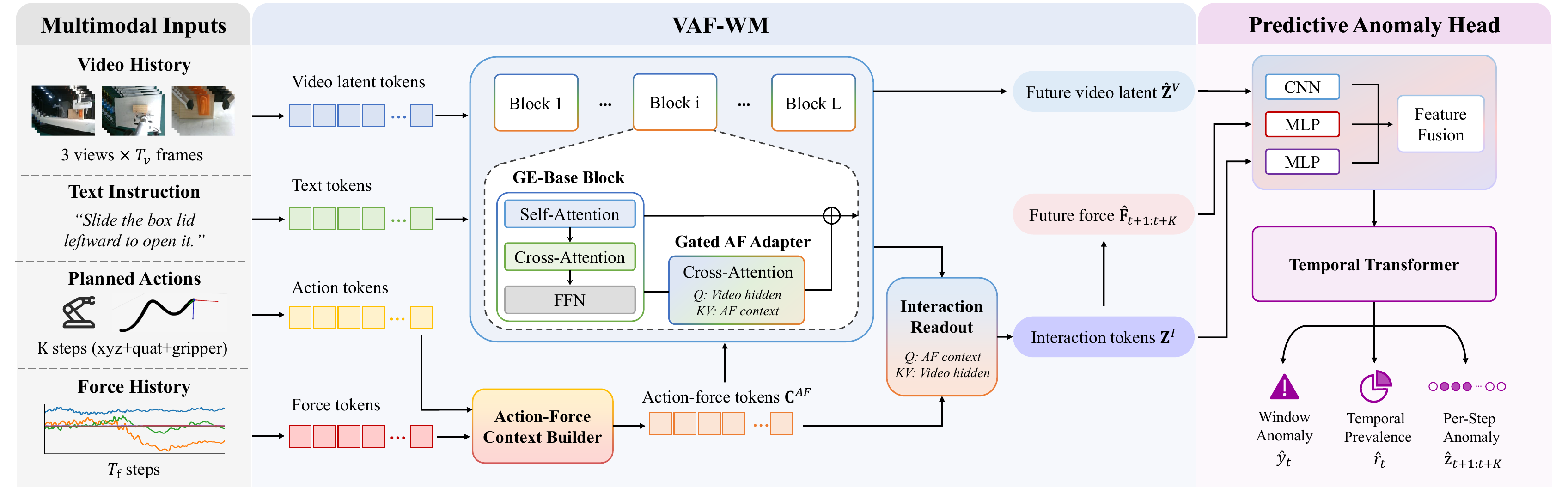}
    \caption{\textbf{MAIA Overview.} The framework consists of a Vision-Action-Force World Model (VAF-WM) and a predictive anomaly head. VAF-WM extends GE-Base with gated action-force adapters to forecast future visual and force dynamics. The anomaly head fuses these predictions with interaction tokens and uses a temporal transformer to predict anomaly occurrence, temporal prevalence, and localization within the future window.}
    \label{fig:model_arch}
    \vspace{-4mm}%
\end{figure*}

\section{Method}
% As illustrated in Fig.~\ref{fig:model_arch}, our framework anticipates future manipulation anomalies by modeling how the robot–environment interaction is expected to evolve under the policy’s planned actions. 
% The Vision-Action-Force World Model (VAF-WM) conditions a pretrained video world model on recent visual and force observations, the task instruction, and planned future actions to roll out future multimodal states. A lightweight predictive anomaly head then aggregates these  predictive signals to estimate future anomaly occurrence and temporal prevalence.
% The framework is trained in three progressive stages: we first adapt the VAF-WM on nominal executions to learn expected interaction dynamics, then train the anomaly head to recognize deviations from the learned dynamics, and finally transfer the framework to novel tasks with limited task-specific data.
As illustrated in Fig.~\ref{fig:model_arch}, MAIA anticipates manipulation anomalies by modeling how robot–environment interactions evolve under the policy’s planned actions. The Vision–Action–Force World Model (VAF-WM) conditions a pretrained video world model on recent visual and force observations, the task instruction, and planned future actions to predict future multimodal states. A lightweight anomaly head then aggregates these predictive signals to estimate future anomaly occurrence and temporal prevalence. MAIA is trained in three stages: nominal interaction dynamics learning, predictive anomaly-head training, and limited-data adaptation to novel tasks.

\subsection{Vision-Action-Force World Model}
We initialize VAF-WM from the pretrained GE-Base backbone~\cite{ge-base} and extend it with explicit action and force conditioning for physical interaction modeling. While GE-Base provides a strong prior for future visual prediction, contact-rich manipulation additionally requires modeling the physical consequences of planned robot motions under the current interaction state. We therefore augment the video world model with an action–force pathway that conditions its predictive rollout on recent force history and future policy actions. In our implementation, the prediction horizon is set to $K=50$ control steps, with dense action and force sequences and sparsely sampled multi-view visual observations.

% \subsubsection{Action–force Context Construction}
\textbf{Action–force Context Construction.}
The future action sequence is encoded into action tokens, while a force-history encoder extracts both temporal variations and a global contact summary from recent physical responses. Each action token queries the temporal force representation through cross-attention, while the global summary modulates the resulting action-aligned features, producing action-force context tokens $\mathbf{C}^{AF} \in \mathbb{R}^{K \times d}$.
This explicit coupling grounds each future action in the encoded physical feedback, allowing the world model to reason about action-dependent physical consequences during rollout.

% By querying the force representation with the action tokens, the model produces action-force context tokens $\mathbf{C}^{AF} \in \mathbb{R}^{K \times d}$.

% \subsubsection{Conditioning the Video Backbone}
\textbf{Conditioning the Video Backbone.}
% We incorporate the action-force context into the GE-Base video transformer through gated cross-attention adapters, allowing the visual representation to retrieve action-conditioned physical feedback. The gated residual design preserves the pretrained representation while integrating the retrieved physical information. The conditioned backbone then predicts future multi-view video latents using the original flow-matching formulation.
We inject the action-force context into selected GE-Base transformer blocks through gated cross-attention adapters. At each adapted block, the video hidden states produced by the transformer block query $\mathbf{C}^{AF}$, and the gated cross-attention output is added back through a residual connection. This preserves the visual dynamics prior inherited from GE-Base while augmenting it with action-conditioned physical information.

% \subsubsection{Interaction Readout and Multimodal Prediction}
\textbf{Interaction Readout and Multimodal Prediction.}
After the conditioned video prediction, the final video hidden states are projected to the predicted future video latents $\hat{\mathbf{Z}}^{V}$. In parallel, the action-force context $\mathbf{C}^{AF}$ queries the final video hidden states through an interaction readout module, producing interaction tokens $\mathbf{Z}^{I} \in \mathbb{R}^{K \times d}$. These tokens associate the planned actions and encoded physical feedback with the predicted visual evolution. A temporal force decoder further maps $\mathbf{Z}^{I}$ to the future force sequence $\hat{\mathbf{F}}_{t+1:t+K}$. 

% \subsubsection{Nominal Dynamics Learning}
During the first training stage, VAF-WM is trained exclusively on nominal executions to learn the expected interaction dynamics. We retain the native flow-matching objective for video prediction and jointly supervise future force prediction,
\begin{equation}
   {L}_{WM} = \lambda_v{L}_{video} + \lambda_f{L}_{force}.
\end{equation}
Here, ${L}_{video} $ denotes the flow-matching loss over the future video latents, while ${L}_{force}$ combines future-force reconstruction with temporal-difference and smoothness regularization.

\subsection{Predictive Anomaly Head}
The predictive anomaly head receives three predictive signals $\{\hat{\mathbf{Z}}^{V}, \mathbf{Z}^{I}, \hat{\mathbf{F}}\}$ from VAF-WM. Dedicated encoders project the inputs into a shared feature space. The temporally sparse visual features are interpolated to align with the $K$-step interaction and force features before fusion, while a global summary of the force history provides auxiliary conditioning. A lightweight temporal transformer then models dependencies across the future horizon. The resulting features are passed to separate heads that estimate the window-level anomaly probability $\hat{y}_t$ and temporal prevalence $\hat{r}_t$, while an auxiliary step head predicts $\hat{z}_{t+1:t+K}$ for step-level supervision.

During the second stage, we freeze VAF-WM and run it in inference mode to generate predictive rollouts from both nominal and anomalous executions for training the anomaly head. This decouples anomaly supervision from nominal-dynamics learning and aligns the head with the predictive representations encountered at deployment. 
The training objective is
\begin{equation}
    {L}_{\mathrm{AH}} = \lambda_y{L}_{\mathrm{window}} + \lambda_p{L}_{\mathrm{prev}} + \lambda_a{L}_{\mathrm{step}},
\end{equation}
where ${L}_{\mathrm{window}}$ and ${L}_{\mathrm{step}}$ are binary cross-entropy losses, and ${L}_{\mathrm{prev}}$ is a smooth $L_1$ loss.

\subsection{Limited-Data Novel-Task Adaptation}
Although VAF-WM captures shared interaction patterns across training tasks, a novel task may introduce shifts in object geometry, manipulation constraints, and physical responses. To adapt with limited task-specific data, we freeze the video backbone learned in Stage 1 and jointly fine-tune the action-force pathway and anomaly head using the following objective:
\begin{equation}
    {L}_{\mathrm{adapt}} = {L}_{\mathrm{AH}} + \lambda_{\mathrm{nom}} {L}_{\mathrm{force}}^{\mathrm{nom}},
\end{equation}
where ${L}_{\mathrm{AH}}$ is the anomaly objective defined above and ${L}_{\mathrm{force}}^{\mathrm{nom}}$ is a force-prediction loss computed exclusively on nominal executions. This regularization anchors the adapted action-force pathway to nominal interaction dynamics while allowing adaptation to task-specific anomaly patterns.

%% file: sections/6_experiments.tex
\begin{table*}[t]
\vspace{5pt}
\centering
\caption{Anomaly anticipation performance on seen and novel tasks, macro-averaged across tasks.}
\vspace{-2mm}
\label{tab:offline_main}
\begin{tabular}{lcccccccc}
\toprule
\multirow{2}{*}{Method} & \multicolumn{4}{c}{Seen Tasks} & \multicolumn{4}{c}{Novel Tasks} \\
\cmidrule(lr){2-5} \cmidrule(lr){6-9}
& AUPRC$\uparrow$ & bACC$\uparrow$ & AR$\uparrow$ &  Med. ALT$\uparrow$ & AUPRC$\uparrow$ & bACC$\uparrow$ & AR$\uparrow$ & Med. ALT$\uparrow$ \\
\midrule
PCA-KMeans~\cite{pca}    & 0.488 & 0.576 & 0.113 & 14.3 & 0.589 & 0.508 & 0.142 & 16.4 \\
LogpZO~\cite{faildetect}        & 0.551 & 0.619 & 0.145 & 16.0 & 0.556 & 0.504 & 0.161 & 24.6 \\
SAFE-LSTM~\cite{SAFE}     & 0.583 & 0.724 & 0.182 & 18.8 & 0.426 & 0.495 & 0.103 & 22.2 \\
Robo-Dopamine~\cite{robo-dopamine} & 0.411 & 0.523 & 0.352 & 18.9 & 0.500 & 0.546 & 0.361 & 17.4 \\
Robometer~\cite{robometer}     & 0.356 & 0.512 & 0.320 & 12.0 & 0.469 & 0.524 & 0.377 & 15.3 \\
WCM~\cite{WCM}           & 0.669 & 0.702 & 0.242 & 17.7 & 0.519 & 0.539 & 0.219 & 14.3 \\
\midrule
\textbf{MAIA(Ours)}  & \textbf{0.738} & \textbf{0.765} & \textbf{0.464} & \textbf{28.0} & \textbf{0.728} & \textbf{0.710} & \textbf{0.408} & \textbf{25.1}\\
\bottomrule
\end{tabular}
\vspace{-2pt}%
\end{table*}

\begin{table*}[t]
\centering
\caption{Ablation of visual, action, and force modeling across task categories. 
% We report bACC on seen and novel tasks.
}
\vspace{-2mm}
\label{tab:modality_ablation}
\begin{tabular}{lcccccccccccc}
\toprule
\multirow{2}{*}{Variant} & \multicolumn{2}{c}{Insertion} & \multicolumn{2}{c}{Wiping} & \multicolumn{2}{c}{Pressing} & \multicolumn{2}{c}{Path-Constrained} & \multicolumn{2}{c}{Articulated-Object} & \multicolumn{2}{c}{Average} \\
\cmidrule(lr){2-3} \cmidrule(lr){4-5} \cmidrule(lr){6-7} \cmidrule(lr){8-9} \cmidrule(lr){10-11} \cmidrule(lr){12-13}
& Seen & Novel & Seen & Novel & Seen & Novel & Seen & Novel & Seen & Novel & Seen & Novel \\
\midrule
MAIA-V   & 0.709 & 0.766 & 0.684 & 0.615  & 0.655 & \textbf{0.668} &  0.724 & 0.654 & 0.674 & 0.624 & 0.689 & 0.665 \\
MAIA-VA  & 0.764 & 0.815 & 0.604 & 0.616  & 0.663 & 0.615 & 0.669 & 0.558 & 0.623 & \textbf{0.637} & 0.665 & 0.648 \\
MAIA & \textbf{0.868} & \textbf{0.832} & \textbf{0.743} & \textbf{0.683} & \textbf{0.702} & 0.662 &  \textbf{0.763} & \textbf{0.742} & \textbf{0.748} & 0.631 & \textbf{0.765} & \textbf{0.710} \\

\bottomrule
\end{tabular}
\vspace{-7pt}%

\end{table*}

\section{Experiments}

\subsection{Experimental Setup}

\subsubsection{Task Splits and Temporal Sampling}
For each task, we partition the trajectories into 100 for training, 20 for validation, and 30 for testing. We use the full training splits of 13 seen tasks for main training and hold out 5 tasks for novel-task evaluation. For the novel-task adaptation, we use 10 training trajectories from each novel task. 
All multimodal streams are aligned to a 30 Hz control timeline. Following GE-Base, each sample spans 128 control steps, with the final $K=50$ steps as the prediction horizon and the preceding steps as interaction history. Visual observations are sparsely sampled, while action and force signals retain the dense control-rate sequence. Windows are extracted every 10 steps, with anomaly targets derived from frame-level annotations within the future horizon.

\subsubsection{Baselines}

We compare against representative methods from four related paradigms. 
(1) \textbf{OOD-based methods}, PCA-KMeans~\cite{pca} and LogpZO~\cite{faildetect}, detect deviations from nominal execution in learned embedding spaces. We fit them per task using nominal segments from successful training trajectories and calibrate a one-sided conformal threshold on nominal validation data.
(2) \textbf{VLA-assisted failure detection}, SAFE-LSTM~\cite{SAFE}, performs supervised sequence monitoring over $\pi_{0.5}$ latent representations. It is trained with our frame-level anomaly labels on seen tasks and evaluated zero-shot on novel tasks, using the same conformal calibration.
(3) \textbf{Reward models}, Robo-Dopamine~\cite{robo-dopamine} and Robometer~\cite{robometer}, are evaluated zero-shot using their pretrained progress estimates. We derive anomaly decisions from relative progress for Robo-Dopamine and from local progress drops and low progress-time correlation for Robometer.
(4) \textbf{World-model-based methods}, WCM~\cite{WCM}, predict execution values and future latent states from interaction history. We retain its original prediction architecture while augmenting reward supervision with our frame-level anomaly labels.
When task-specific fitting or calibration is required, only the corresponding training and validation splits are used.

\subsubsection{Metrics}
We evaluate the methods from two perspectives. At the window level, we report AUPRC and balanced accuracy (bACC) over windows after interaction onset. For threshold-dependent metrics, we threshold our window-level anomaly probability $\hat{y}_t$ at 0.5, while baselines follow the method-specific decision rules described above. We also report the mean absolute error (MAE) of our temporal anomaly prevalence predictions. At the event level, we evaluate annotated anomaly segments using Anticipation Rate (AR) and Anticipation Lead Time (ALT). AR is the fraction of events for which a sequence of positive predictions begins within the 50 steps preceding anomaly onset and persists until onset. ALT measures the number of steps from the beginning of this sustained warning to anomaly onset, with the median reported over successfully anticipated events.

\begin{figure}[t]
    \centering
    \includegraphics[width=\linewidth]{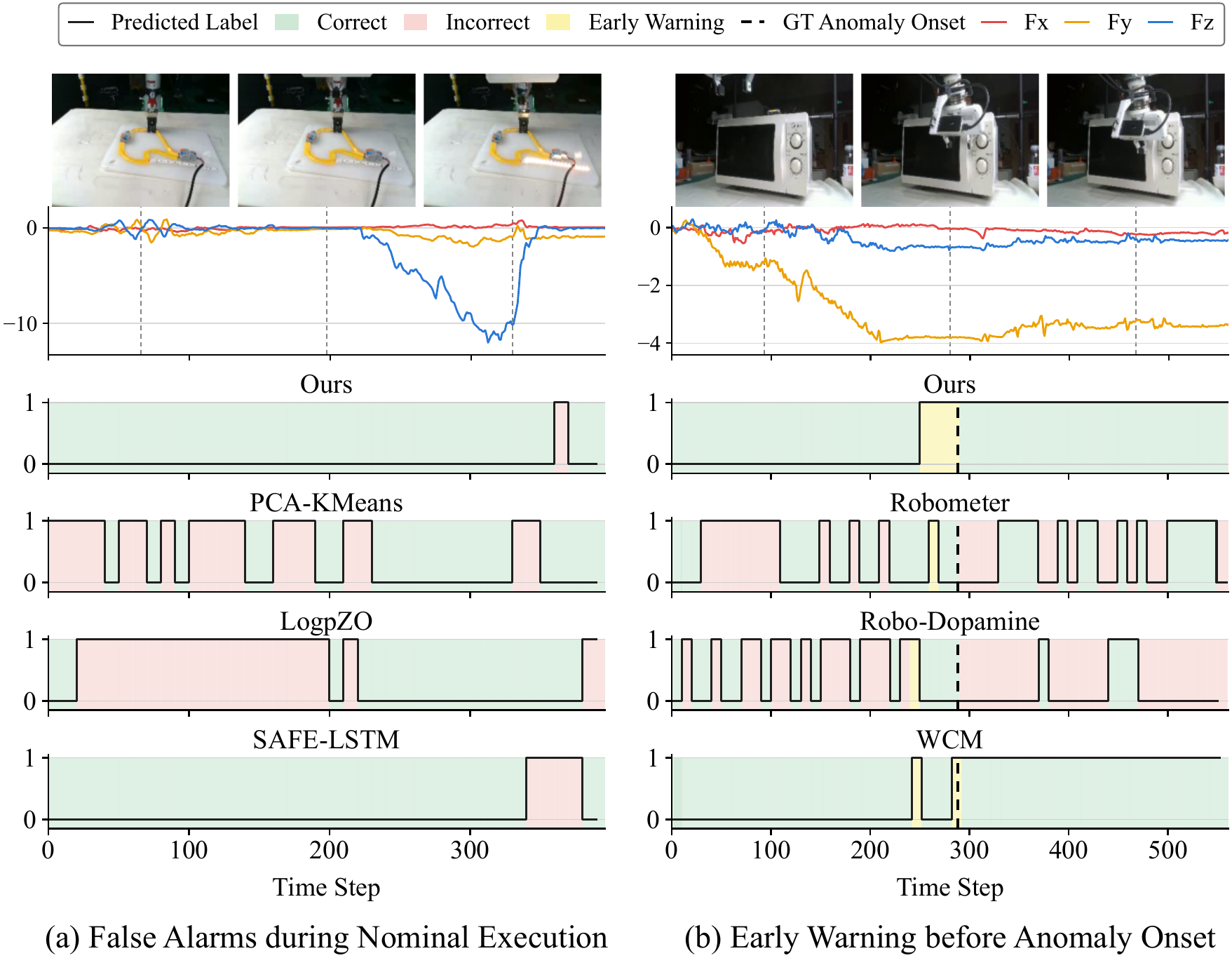}
    \caption{Qualitative comparison of anomaly predictions. Shaded regions indicate correct predictions (green), prediction errors (red), and early warnings (yellow) prior to the true anomaly onset (dashed line). (a) Nominal execution: our method produces fewer false alarms than the baselines. (b) Anomalous execution: our method provides an earlier sustained warning than WCM, while reward-model warnings are intermittent.}
    \label{fig:baseline_cases}
    \vspace{-13pt}%
\end{figure}

\subsection{Comparative Evaluation}
Table~\ref{tab:offline_main} compares offline anomaly prediction performance on seen and novel tasks. Our method achieves the strongest overall performance across both task splits.

% \textbf{Anomaly Discrimination.} OOD-based methods are effective at identifying departures from nominal feature distributions, but such deviations do not necessarily correspond to anomalous interaction, as they may also arise from benign differences in execution speed, contact timing, object pose, or intermediate visual states.
% SAFE-LSTM improves upon this by exploiting VLA latent features and supervised frame-level anomaly labels. However, its bACC drops from 0.724 to 0.495 under zero-shot transfer, indicating limited generalization of the anomaly cues across contact-rich manipulation tasks. 
% % In contrast, our method does not rely solely on feature-space deviation or semantic representations. By explicitly modeling the evolution of visual and force responses under planned actions, VAF-WM provides a physically grounded basis for identifying anomalous interactions, leading to consistently strong discrimination across both task splits.
% In contrast, our method goes beyond feature-space deviations and semantic representations by explicitly modeling how visual and force responses evolve under planned actions. 
% % The resulting predictive evidence provides a physically grounded basis for identifying anomalous interactions, as illustrated by the reduced false alarms in Fig.~\ref{fig:baseline_cases}(a).
% The resulting predictive evidence provides a physically grounded basis for identifying anomalous interactions, as illustrated by the reduced false alarms during a representative nominal execution in Fig.~\ref{fig:baseline_cases}(a).

\textbf{Anomaly Discrimination.} OOD-based methods are effective at identifying departures from nominal feature distributions, but such deviations do not necessarily indicate anomalous interactions, as they may arise from benign variations in execution speed, contact timing, object pose, or intermediate visual states.
SAFE-LSTM improves upon this by exploiting VLA latent features and supervised frame-level anomaly labels. However, its bACC drops from 0.724 to 0.495 under zero-shot transfer, indicating limited cross-task generalization.
In contrast, our method explicitly models how visual and force responses evolve under planned actions, providing physically grounded evidence for identifying anomalous interactions. In a representative nominal execution, MAIA produces fewer false alarms than the baselines, as shown in Fig.~\ref{fig:baseline_cases}(a).

\textbf{Anomaly Anticipation.} Reward models provide moderate anticipation even without explicit future prediction, as progress may degrade before anomaly onset. However, task progress is only an indirect indicator of anomalous interaction, resulting in weaker discrimination. This is particularly evident for Robometer, whose relatively high anticipation rate is accompanied by a high false-positive rate.
WCM is more naturally aligned with predictive monitoring, but its one-step dynamics objective provides limited temporal reach, leading to relatively modest anticipation performance. 
Our method instead predicts anomalous interactions directly over the full future horizon, providing a better balance between anomaly discrimination and anticipation, as shown in Fig.~\ref{fig:baseline_cases}(b). We also estimate temporal prevalence, achieving an MAE of 0.18 on seen tasks and 0.28 on novel tasks.

\begin{figure*}[t]
    \centering
    \vspace{5pt}
    \includegraphics[width=\linewidth]{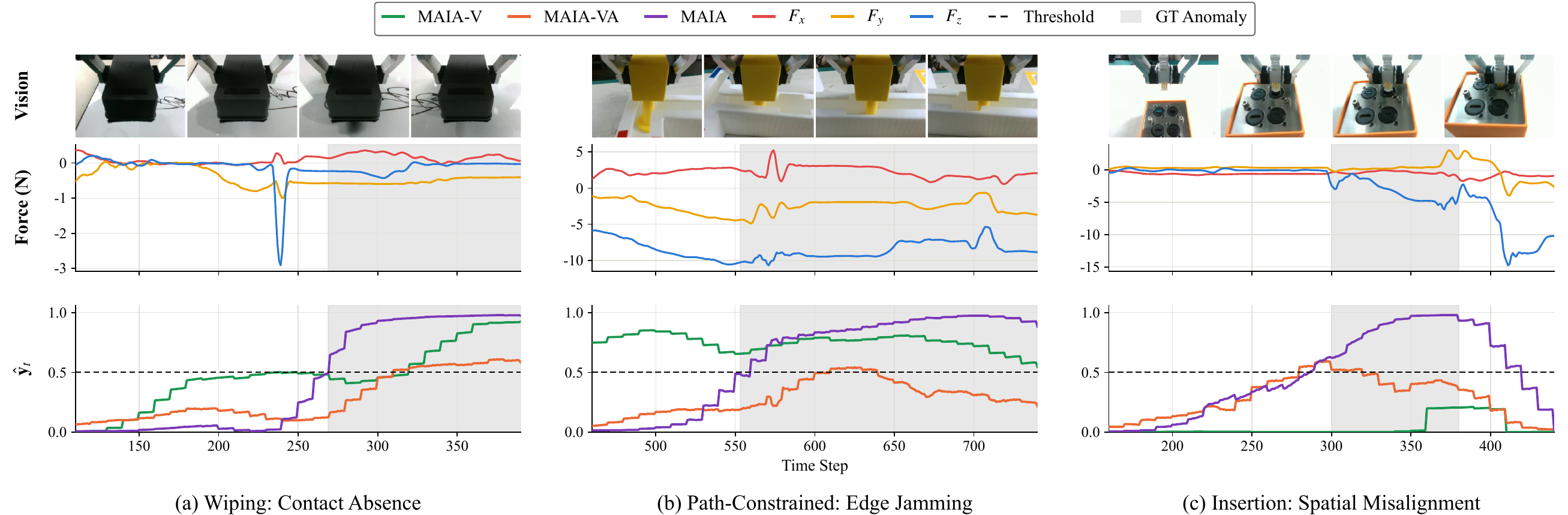}
    \caption{Qualitative modality ablation on representative anomaly cases. The examples illustrate how incorporating physical feedback enables MAIA to distinguish anomalous interactions that remain ambiguous when only visual and action cues are available.}
    \label{fig:ablation_cases}
    \vspace{-7pt}%
\end{figure*}

\begin{figure}[tb]
    \centering
    \includegraphics[width=\linewidth]{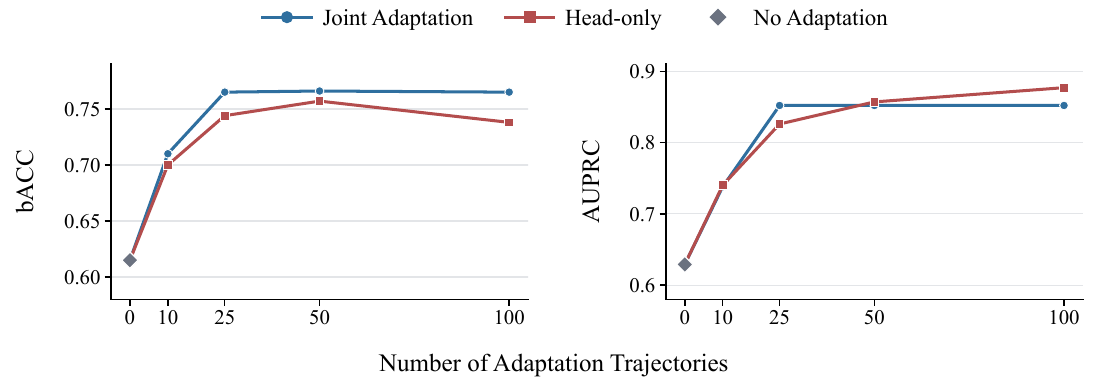}
    \caption{Novel-task adaptation with varying amounts of task-specific data.}
    \label{fig:novel_adaptation}
    \vspace{-4.5mm}%
\end{figure}

\subsection{Ablation Studies}
% To isolate the contributions of planned actions and physical feedback, we construct three modality configurations by progressively removing the corresponding pathways and predictive signals. MAIA-V removes the action-force pathway and feeds only predicted visual latents to the anomaly head. MAIA-VA retains action conditioning but removes force input and force prediction, with the anomaly head receiving predicted visual latents and interaction tokens. MAIA-VAF denotes the full model, where predicted force trajectories are additionally provided to the anomaly head. All variants follow the same training protocol and anomaly supervision.
To isolate the contributions of planned actions and physical feedback, we evaluate MAIA with two variants. MAIA-V removes the action-force pathway and feeds only predicted visual latents to the anomaly head. MAIA-VA retains action conditioning but removes force input and force prediction, with the anomaly head receiving predicted visual latents and interaction tokens. All configurations use the same training protocol and anomaly supervision, 
% and their quantitative comparison is reported in Table~\ref{tab:modality_ablation}.
and are quantitatively compared using bACC in Table~\ref{tab:modality_ablation}.

% The ablation reveals that action conditioning alone does not consistently improve anomaly discrimination, particularly in Wiping and Path-Constrained manipulation, where similar local motions might correspond to different interaction states. Adding force feedback solves this issue, improving bACC over the MAIA-VA setting by 0.139 and 0.094, respectively. 
% As illustrated in Fig.~\ref{fig:ablation_cases}, visually plausible motions can still coincide with subtle interaction changes(e.g., loss of surface contact or unexpected mechanical resistance); the predicted force responses directly expose these changes.
% Insertion shows a complementary pattern, where action information is already useful and force provides further gains by distinguishing geometrically plausible motions from invalid physical contact.
% Overall, the results show that grounding planned actions in physical interaction cues substantially improves anomaly prediction, particularly in contact regimes that are difficult to disambiguate from vision alone.
Action conditioning alone does not consistently improve anomaly discrimination, particularly in Wiping and Path-Constrained manipulation, where similar local motions may correspond to different interaction states. Adding force feedback solves this issue, improving bACC over MAIA-VA by 0.139 and 0.094, respectively.
As illustrated in Fig.~\ref{fig:ablation_cases}, visually plausible motions may still involve subtle interaction changes (e.g., loss of surface contact or unexpected mechanical resistance), which are revealed by the predicted force responses.
Insertion shows a complementary pattern, where action information is already useful and force further improves discrimination between geometrically plausible motions and invalid physical contact.
Overall, the results show that grounding planned actions in physical interaction cues improves anomaly prediction, particularly in contact regimes that are difficult to disambiguate from vision alone.

\subsection{Novel-Task Adaptation}
We compare joint adaptation with head-only adaptation on novel tasks, varying the amount of task-specific data from 0 to 100 trajectories. As shown in Fig.~\ref{fig:novel_adaptation}, the model trained on seen tasks captures transferable interaction dynamics across contact-rich manipulation tasks, achieving a bACC of 0.615 on novel tasks without task-specific tuning.
With limited data, joint adaptation improves rapidly and largely saturates after 25 trajectories, reaching a bACC of 0.762. In contrast, head-only adaptation is less data-efficient and yields lower performance in the low-data regime. These results suggest that adapting the interaction representation is more effective than adjusting the anomaly head alone for novel-task transfer.

\subsection{Real-Robot Policy Integration}
We evaluate whether the predicted anomaly probability $\hat y_t$ can guide action selection during real-robot deployment. At each planning step, a frozen $\pi_{0.5}$ policy generates 8 candidate action chunks. MAIA estimates the anomaly probability over the prediction horizon for each chunk and selects the one with the lowest probability. We execute the first 20 actions of the selected chunk while asynchronously computing the next plan. We use the trained model for seen tasks and its adapted variants for novel tasks. We compare this strategy with the original policy on all 18 tasks across 5 categories, with 30 trials per task for each method. As shown in Table~\ref{tab:real_robot}, anomaly-aware selection improves the overall task success rate from 52.6\% to 61.5\%. Consistent with the ablation, it yields the largest gains in Wiping and Path-Constrained tasks, 
% where physical interaction cues are more informative.
where reliable decision-making depends more strongly on physical interaction cues.

% The largest gains are observed in Wiping and Path-Constrained tasks, while average success in Insertion remains unchanged.

% \begin{table}[t]
% \centering
% \caption{Real-robot task success rates (\%) .}
% \label{tab:real_robot}
% \resizebox{\columnwidth}{!}{
% \begin{tabular}{lcccccc}
% \toprule
% Method & Insertion & Wiping & Pressing & Path. & Articulated & Overall \\
% \midrule
% $\pi_{0.5}$ & \textbf{63.3} & 46.7 & 48.3 & 48.9 & 53.3 & 52.6 \\
% $\pi_{0.5}$ + MAIA & \textbf{63.3} & \textbf{66.7}  & \textbf{54.2} & \textbf{62.2} & \textbf{62.5} & \textbf{61.5} \\
% \bottomrule
% \end{tabular}}
% \vspace{-5pt}%
% \end{table}

\begin{table}[t]
\centering
% \caption{Real-robot task success rates (\%) with anomaly-aware action selection.}
\caption{Real-robot task success rates (\%).}
\label{tab:real_robot}
\vspace{-3pt}
\setlength{\tabcolsep}{2.5pt}
\begin{tabular}{lcccccc}
\toprule
Method & Insertion & Wiping & Pressing & Path. & Articulated & Overall \\
\midrule
$\pi_{0.5}$ 
& \textbf{63.3} & 46.7 & 48.3 & 48.9 & 53.3 & 52.6 \\
% \makecell[l]{$\pi_{0.5}$ + anomaly \\ selection}
$\pi_{0.5}$ + MAIA
& \textbf{63.3} & \textbf{66.7} & \textbf{54.2}  & \textbf{62.2} & \textbf{62.5} & \textbf{61.5} \\
\bottomrule
\end{tabular}
\vspace{-8pt}%
\end{table}

%% file: sections/7_conclusion.tex
\section{Conclusion and Limitations}
In this work, we formulate anomaly anticipation in contact-rich manipulation as a predictive interaction problem: instead of detecting deviations after they occur, we predict whether planned actions will lead to anomalous physical interactions. We introduce CRIA, a real-world multimodal dataset with fine-grained temporal anomaly annotations, and MAIA, which models the visual and force consequences of planned actions to anticipate anomalies before onset. Experiments on seen and novel tasks show improved anomaly discrimination and earlier warnings. Beyond monitoring, predicted anomaly probabilities can guide action selection and improve real-robot task success without policy retraining. These results demonstrate a shift from reactive failure detection toward proactive safeguarding of robot execution.

Despite these results, several limitations remain. First, finite-horizon forecasting and fixed-length history may limit prediction reliability over longer horizons and for anomalies requiring extended interaction context. Second, experiments on a single robot platform and fixed force/torque sensing setup leave cross-embodiment and cross-sensor generalization unexplored. Finally, the framework focuses on anomaly anticipation rather than root-cause diagnosis or autonomous recovery, motivating future work on closed-loop control.

%% file: sections/X_appendix.tex
\clearpage
\appendix
% % 

% \appendices
% \section{Dataset}
% \section{Additional Experiments}
% \section{Implementation Details}

\subsection{Dataset}
\subsubsection{Dataset Statistics}

CRIA contains 2,852 trajectories spanning 18 contact-rich manipulation tasks across five interaction categories: Insertion, Wiping, Pressing, Path-Constrained Manipulation, and Articulated-Object Manipulation. Table~\ref{tab:dataset_statistics} summarizes the number of trajectories, anomalous trajectories, anomaly events, and anomalous frames for each task. An anomalous trajectory is a trajectory containing at least one anomalous segment, while an anomaly event corresponds to an individual anomalous segment. The anomalous-frame ratio measures the fraction of frames labeled as anomalous over all frames. The 2,328 anomaly events across 1,817 anomalous trajectories show that some trajectories contain multiple temporally distinct anomalous segments, highlighting the temporal structure of the collected anomalies.

\subsubsection{Task Categories and Anomaly Modes}
We organize the CRIA tasks by the physical interaction regimes that govern execution rather than by task semantics alone. Although tasks within a category may involve different objects and goals, they share common contact constraints and recurring anomaly modes. 

Insertion tasks involve precise geometric alignment under tight spatial tolerances, with successful execution requiring motion along a feasible mating direction. Typical anomalies include misalignment, tilting, partial insertion, excessive contact force, jamming, and repeated retract-and-retry behavior.

Wiping tasks require sustained normal contact during tangential surface motion. Their execution is therefore sensitive to contact stability and force distribution, with anomalies including loss of contact, uneven pressure, trajectory deviation, and incomplete surface coverage.

Pressing tasks are dominated by localized contact and force buildup. Successful execution depends on applying force at the correct position and direction to trigger the intended mechanical response. Typical anomalies include insufficient force, misaligned contact, redundant contact, and incomplete actuation.

Path-constrained manipulation requires the manipulated object or end-effector to follow a prescribed geometric path imposed by the surrounding structure. Violations of these constraints may produce unexpected resistance, with typical anomalies including trajectory deviation and slipping.

Articulated-object manipulation is governed by the kinematic structure of the manipulated object, including prismatic and revolute joints. Feasible motion is restricted to the corresponding articulation path, while typical anomalies include missed engagement, jamming or sticking caused by incorrect force direction, and prolonged static pauses resulting from misaligned grasping.

These interaction characteristics are represented by 18 tasks spanning the five categories and covering distinct contact and motion constraints. Table~\ref{tab:task_details} summarizes the task specifications, including the instruction and a concise interaction description for each task.

\begin{table}[t]
    \centering
    % \begin{threeparttable}
    \caption{Statistics of the CRIA dataset.}
    \label{tab:dataset_statistics}
    \begin{tabular}{lcccc}
        \toprule
        \textbf{Category / Task} &
        \textbf{Traj.} &
        \makecell{\textbf{Anomalous}\\\textbf{Traj.}} &
        \makecell{\textbf{Anomaly}\\\textbf{Events}} &
        \makecell{\textbf{Anomalous}\\\textbf{Frames}} \\
        \midrule
        \multicolumn{5}{l}{\textit{Insertion}} \\
        \quad Insert-Plug-2   & 150 & 128 & 221 & 35.13\% \\
        \quad Insert-Plug-3   & 150 & 133 & 171 & 44.11\% \\
        \quad Insert-Type-C    & 150 & 139 & 156 & 38.83\% \\
        \quad Insert-NI       & 150 & 140 & 161 & 50.30\% \\
        \textbf{Subtotal}     & \textbf{600} & \textbf{540} & \textbf{709} & \textbf{42.11\%} \\
        \midrule
        \multicolumn{5}{l}{\textit{Wiping}} \\
        \quad Wipe-Board      & 199 & 85 & 86  & 10.07\% \\
        \quad Wipe-Bowl       & 150 & 80 & 97  & 14.21\% \\
        \quad Wipe-Vase       & 151 & 95 & 103 & 23.40\% \\
        \textbf{Subtotal}     & \textbf{500} & \textbf{260} & \textbf{286} & \textbf{15.59\%} \\
        \midrule
        \multicolumn{5}{l}{\textit{Pressing}} \\
        \quad Press-Seal      & 150 & 79 & 87  & 13.01\% \\
        \quad Press-Switch-1  & 150 & 68 & 72  & 22.76\% \\
        \quad Press-Switch-2  & 150 & 89 & 121 & 16.99\% \\
        \quad Press-Bottle    & 150 & 94 & 138 & 32.70\% \\
        \textbf{Subtotal}     & \textbf{600} & \textbf{330} & \textbf{418} & \textbf{21.57\%} \\
        \midrule
        \multicolumn{5}{l}{\textit{Path-Constrained Manipulation}} \\
        \quad Push-I          & 180 & 161 & 257 & 16.85\% \\
        \quad Zipper          & 181 & 75  & 90  & 7.38\% \\
        \quad Push-C          & 180 & 149 & 212 & 14.84\% \\
        \textbf{Subtotal}     & \textbf{541} & \textbf{385} & \textbf{559} & \textbf{13.28\%} \\
        \midrule
        \multicolumn{5}{l}{\textit{Articulated-Object Manipulation}} \\
        \quad Microwave       & 151 & 79 & 89  & 16.32\% \\
        \quad Open-Box-1      & 150 & 65 & 71  & 11.67\% \\
        \quad Open-Box-2      & 160 & 72 & 83  & 14.39\% \\
        \quad Pull-Drawer     & 150 & 86 & 113 & 13.47\% \\
        \textbf{Subtotal}     & \textbf{611} & \textbf{302} & \textbf{356} & \textbf{14.04\%} \\
        \midrule
        \textbf{Total}        & \textbf{2852} & \textbf{1817} & \textbf{2328} & \textbf{20.88\%} \\
        \bottomrule
    \end{tabular}
    % }
    % \begin{tablenotes}
    % \footnotesize
    % \item ``Anomalous Traj.'' denotes trajectories containing at least one anomalous segment.
    % \item ``Anomaly Events'' denotes the number of anomaly segments.
    % \item ``Anomalous Frames'' denotes the percentage of anomalous frames.
    % \end{tablenotes}
    % \end{threeparttable}
\end{table}

\begin{table*}[t]
    \centering
    \caption{Task specifications in CRIA.}
    \label{tab:task_details}
    \setlength{\tabcolsep}{6pt}
    \renewcommand{\arraystretch}{0.95}

    \begin{tabular}{@{}p{4.25cm}p{4.25cm}p{4.25cm}p{4.25cm}@{}}
       \toprule

        \multicolumn{4}{l}{\textit{Insertion}}   \\[3pt]

        \begin{minipage}[t]{4.45cm}
            \centering
            \includegraphics[width=2.0cm]{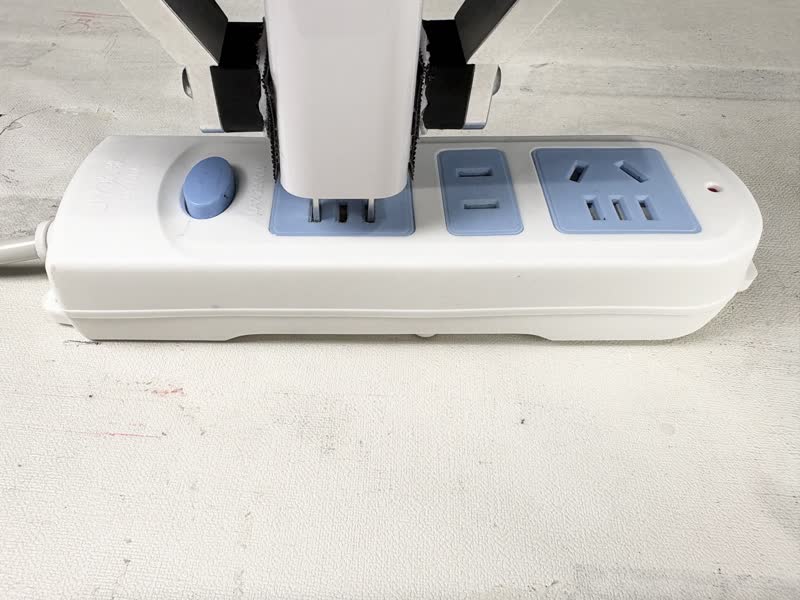}\\[3pt]
            \raggedright
            \textbf{Task:} Insert-Plug-2\\  
            \textbf{Instruction:} Insert a two-prong plug.\\  
            \textbf{Description:} Insert the two-prong power plug into its socket with precise alignment.
        \end{minipage}
        &
        \begin{minipage}[t]{4.45cm}
            \centering
            \includegraphics[width=2.0cm]{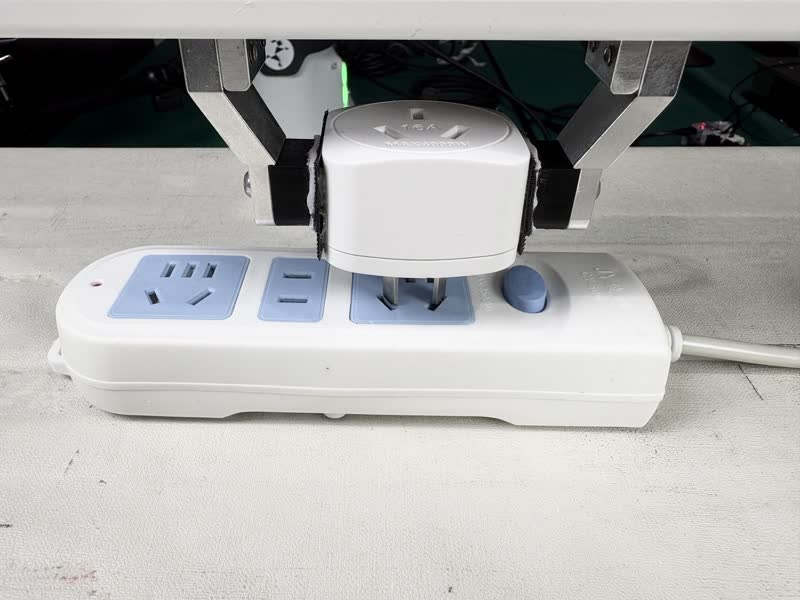}\\[3pt]
            \raggedright
            \textbf{Task:} Insert-Plug-3\\  
            \textbf{Instruction:} Insert a three-prong plug.\\  
            \textbf{Description:} Insert the three-prong power plug into its socket with precise alignment.
        \end{minipage}
        &
        \begin{minipage}[t]{4.45cm}
            \centering
            \includegraphics[width=2.0cm]{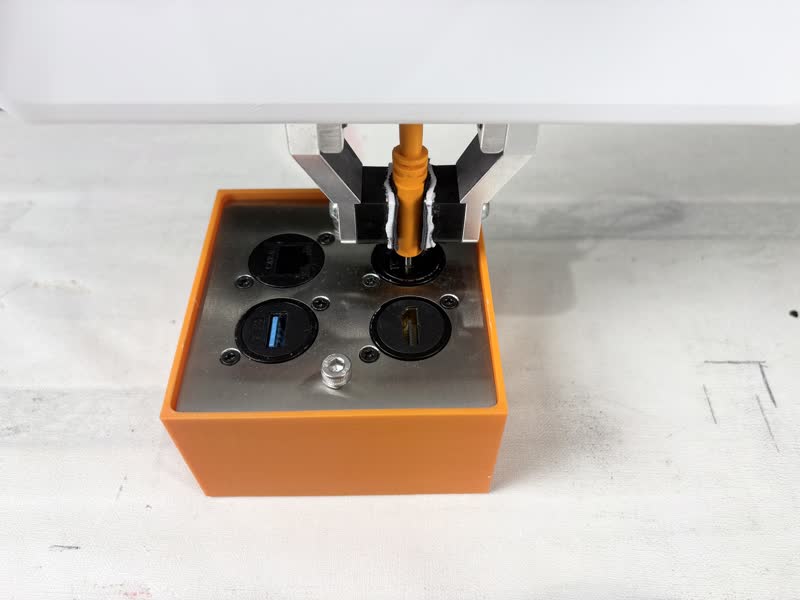}\\[3pt]
            \raggedright
            \textbf{Task:} Insert-Type-C\\  
            \textbf{Instruction:} Insert the typec.\\  
            \textbf{Description:} Insert the Type-C connector into its port with correct orientation and fine alignment.
        \end{minipage}
        &
        \begin{minipage}[t]{4.45cm}
            \centering
            \includegraphics[width=2.0cm]{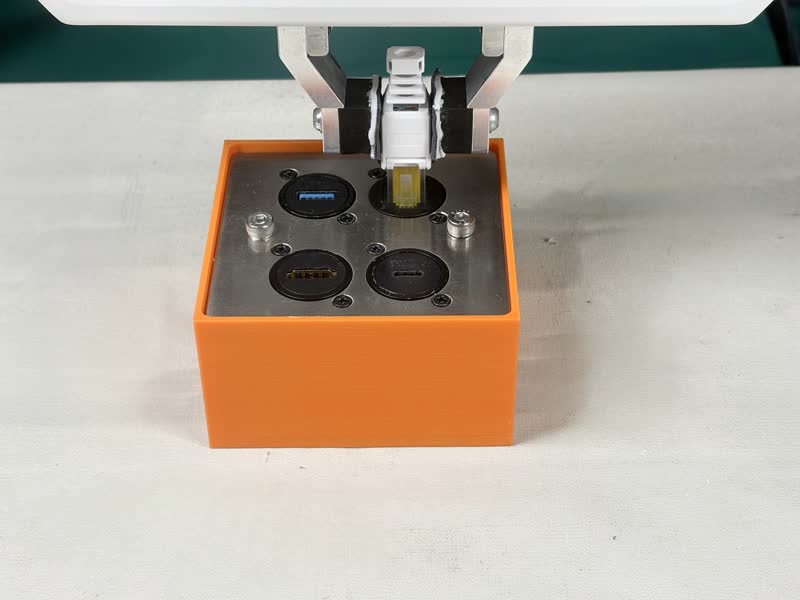}\\[3pt]
            \raggedright
            \textbf{Task:} Insert-NI\\  
            \textbf{Instruction:} Insert the ethernet port.\\  
            \textbf{Description:} Insert the RJ45 plug into the ethernet port with precise alignment.
        \end{minipage}
        \\[4pt]

        \midrule

        \multicolumn{4}{l}{\textit{Wiping}} \\[3pt]

        \begin{minipage}[t]{4.45cm}
            \centering
            \includegraphics[width=2.0cm]{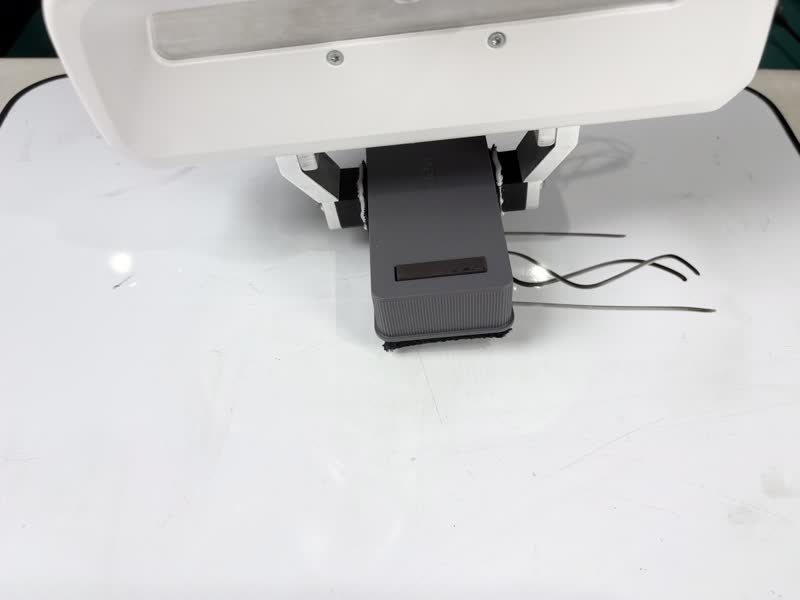}\\[3pt]
            \raggedright
            \textbf{Task:} Wipe-Board\\  
            \textbf{Instruction:} Erase the whiteboard.\\  
            \textbf{Description:} Bring the eraser into contact with the whiteboard, then move it across the board along a tangential wiping path while maintaining normal contact to remove the markings.
        \end{minipage}
        &
        \begin{minipage}[t]{4.45cm}
            \centering
            \includegraphics[width=2.0cm]{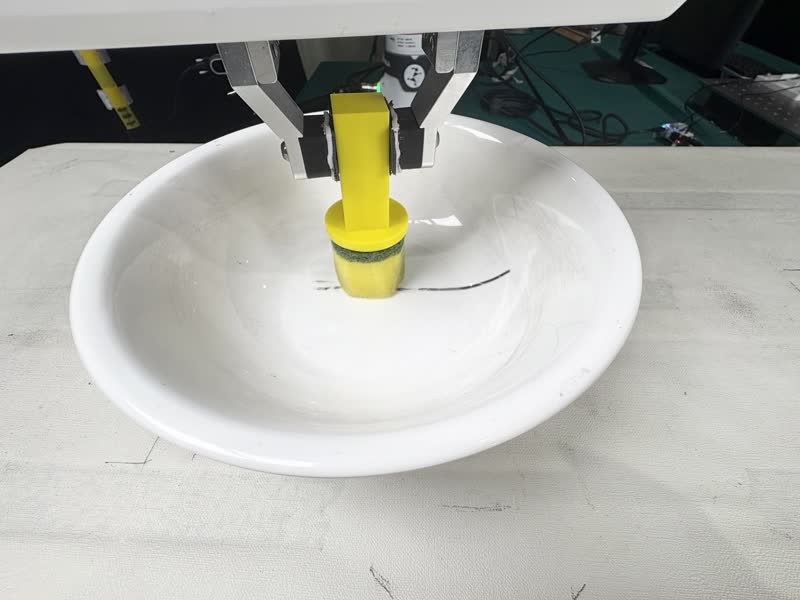}\\[3pt]
            \raggedright
            \textbf{Task:} Wipe-Bowl\\  
            \textbf{Instruction:} Wipe the bowl from left to right.\\  
            \textbf{Description:} Use the gripper-held wiping tool to contact the bowl’s inner surface, then move it along the inner rim while maintaining normal contact to clean the bowl.
        \end{minipage}
        &
        \begin{minipage}[t]{4.45cm}
            \centering
            \includegraphics[width=2.0cm]{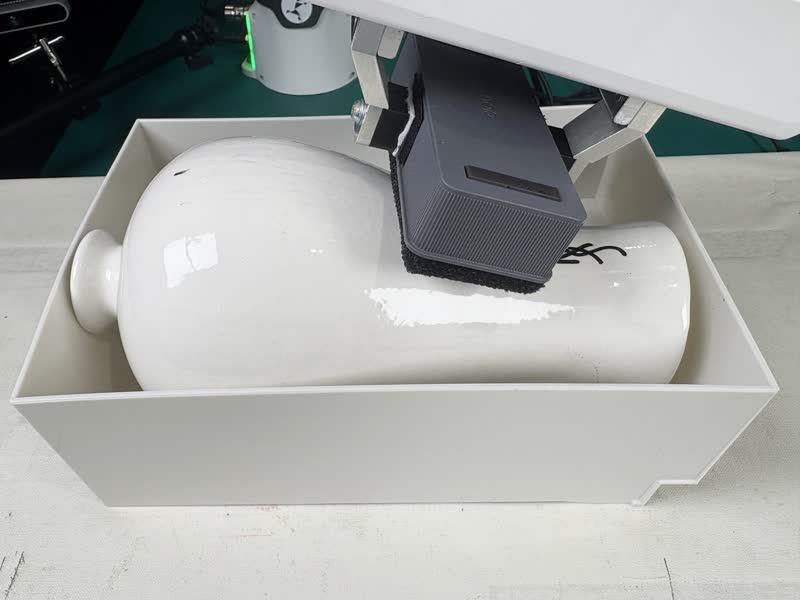}\\[3pt]
            \raggedright
            \textbf{Task:} Wipe-Vase\\  
            \textbf{Instruction:} Wipe the vase.\\  
            \textbf{Description:} Bring the eraser into contact with the vase’s outer surface, then move it along the vase’s outer contour while maintaining normal contact to remove the markings.
        \end{minipage}
        &
        \begin{minipage}[t]{4.45cm}
        \end{minipage}
        \\[4pt]

        \midrule

        \multicolumn{4}{l}{\textit{Pressing}} \\[3pt]

        \begin{minipage}[t]{4.45cm}
            \centering
            \includegraphics[width=2.0cm]{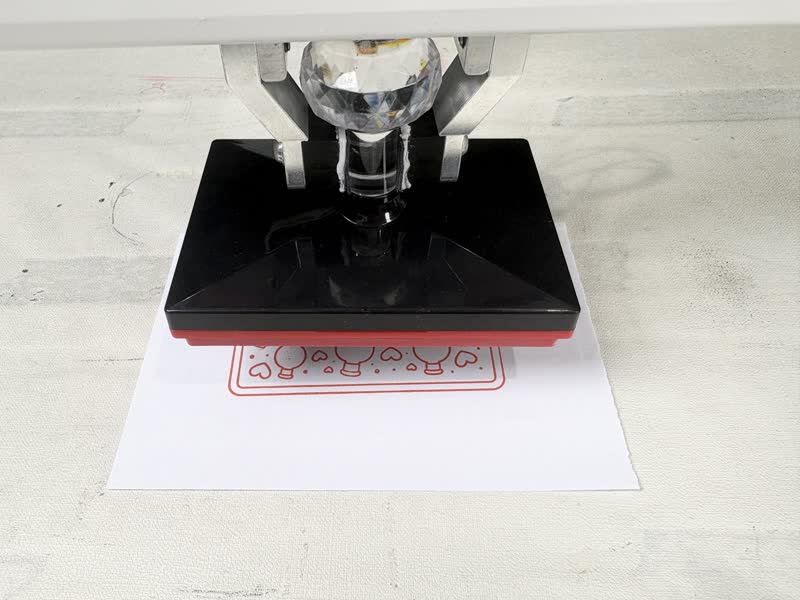}\\[3pt]
            \raggedright
            \textbf{Task:} Press-Seal \\[0pt]
            \textbf{Instruction:} Press the seal on the white paper.\\[0pt]
            \textbf{Description:} Press the seal onto the white paper with adequate axial force along the pressing axis to ensure a clear impression.
        \end{minipage}
        &
        \begin{minipage}[t]{4.45cm}
            \centering
            \includegraphics[width=2.0cm]{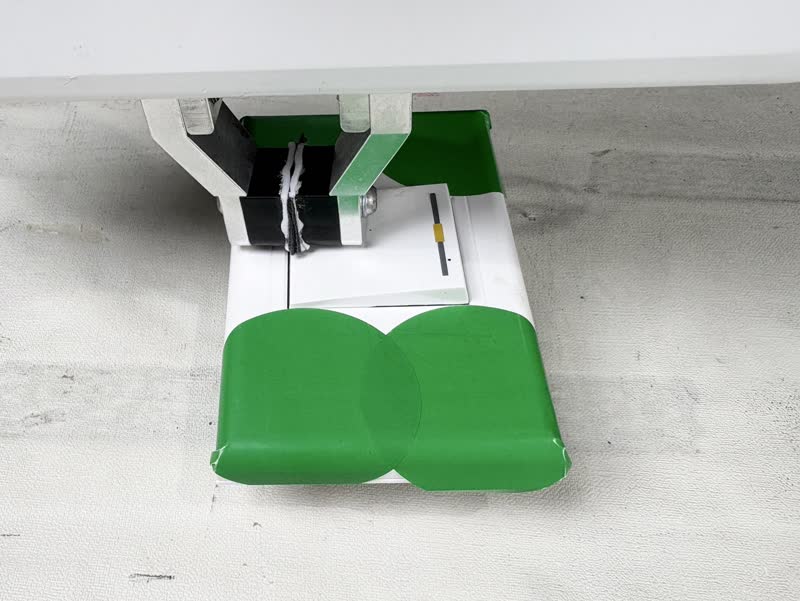}\\[3pt]
            \raggedright
            \textbf{Task:} Press-Switch-1\\
            \textbf{Instruction:} Press the switch.\\
            \textbf{Description:} Contact the rocker switch, then press it with localized force at the correct position and angle to toggle its state.
        \end{minipage}
        &
        \begin{minipage}[t]{4.45cm}
            \centering
            \includegraphics[width=2.0cm]{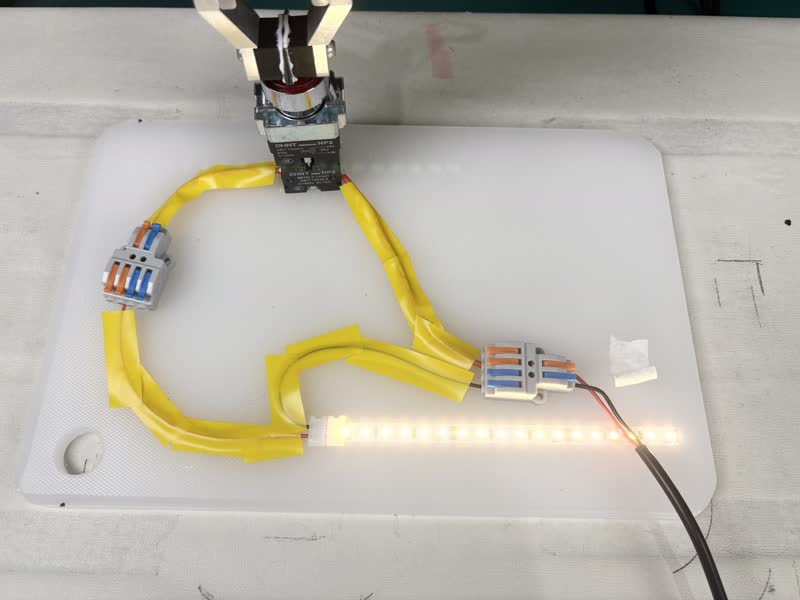}\\[3pt]
            \raggedright
            \textbf{Task:} Press-Switch-2\\
            \textbf{Instruction:} Press the red button.\\
            \textbf{Description:} Contact the red push-button, then press it with sufficient normal force and a short, controlled stroke to light up the LED strip.
        \end{minipage}
        &
        \begin{minipage}[t]{4.45cm}
            \centering
            \includegraphics[width=2.0cm]{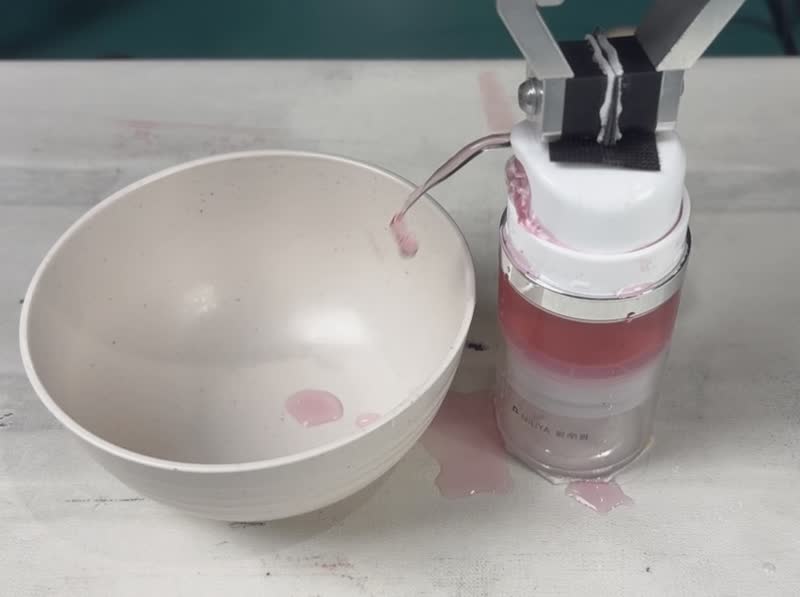}\\[3pt]
            \raggedright
            \textbf{Task:} Press-Bottle\\
            \textbf{Instruction:} Press the pump to dispense liquid.\\
            \textbf{Description:} Contact the pump, then press it with adequate axial force along the pressing axis to press out liquid while keeping the bottle stable.
        \end{minipage}
        \\[4pt]

        \midrule

        \multicolumn{4}{l}{\textit{Path-Constrained Manipulation}} \\[3pt]

        \begin{minipage}[t]{4.45cm}
            \centering
            \includegraphics[width=2.0cm]{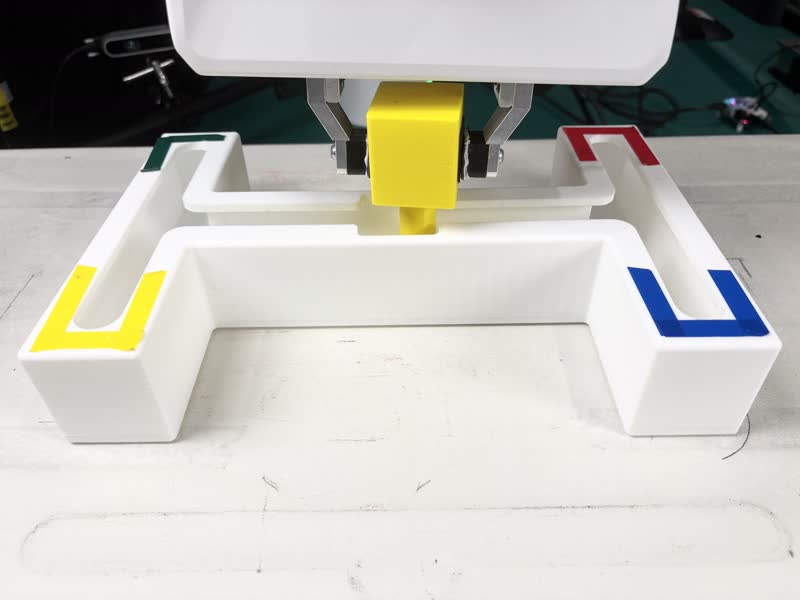}\\[3pt]
            \raggedright
            \textbf{Task:} Push-I\\
            \textbf{Instruction:} Push the block from red area to yellow area.\\
            \textbf{Description:} Grip the block, then push it from left to right along a constrained I-shaped path.
        \end{minipage}
        &
        \begin{minipage}[t]{4.45cm}
            \centering
            \includegraphics[width=2.0cm]{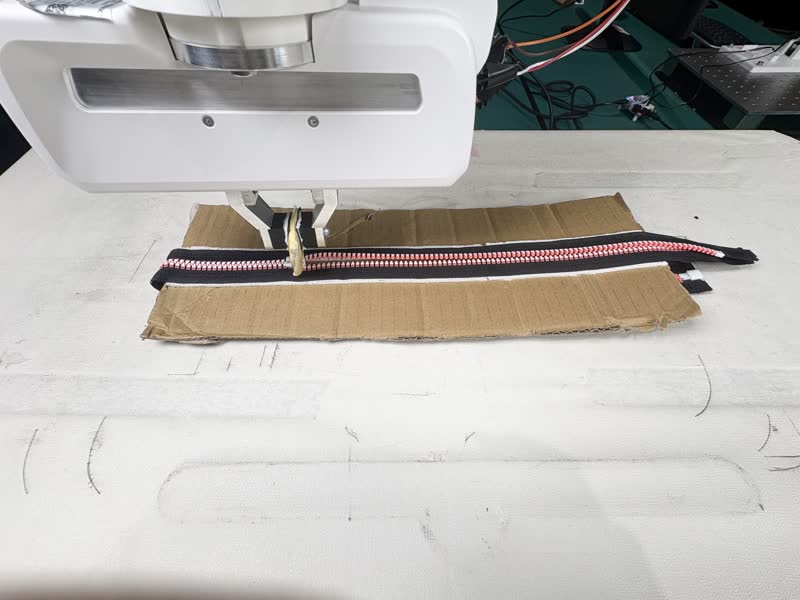}\\[3pt]
            \raggedright
            \textbf{Task:} Zipper\\  
            \textbf{Instruction:} Pull the zipper from left to right.\\  
            \textbf{Description:} Grip the zipper pull tab, then pull it from left to right along its track to open the zipper.
        \end{minipage}
        &
        \begin{minipage}[t]{4.45cm}
            \centering
            \includegraphics[width=2.0cm]{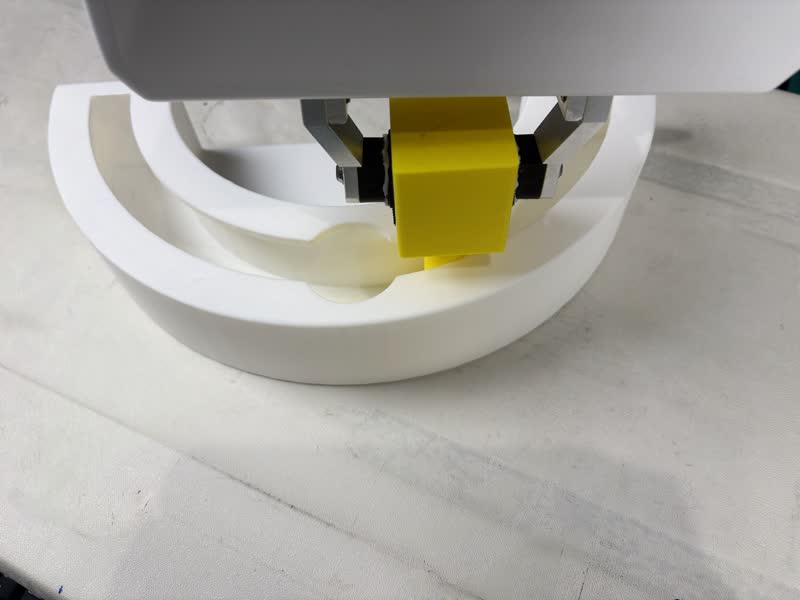}\\[3pt]
            \raggedright
            \textbf{Task:} Push-C\\  
            \textbf{Instruction:} Push the block from left to right.\\  
            \textbf{Description:} Grip the block, then push it from left to right along a constrained C-shaped path.
        \end{minipage}
        &
        \begin{minipage}[t]{4.45cm}
        \end{minipage}
        \\[4pt]

        \midrule

        \multicolumn{4}{l}{\textit{Articulated-Object Manipulation}} \\[3pt]

        \begin{minipage}[t]{4.45cm}
            \centering
            \includegraphics[width=2.0cm]{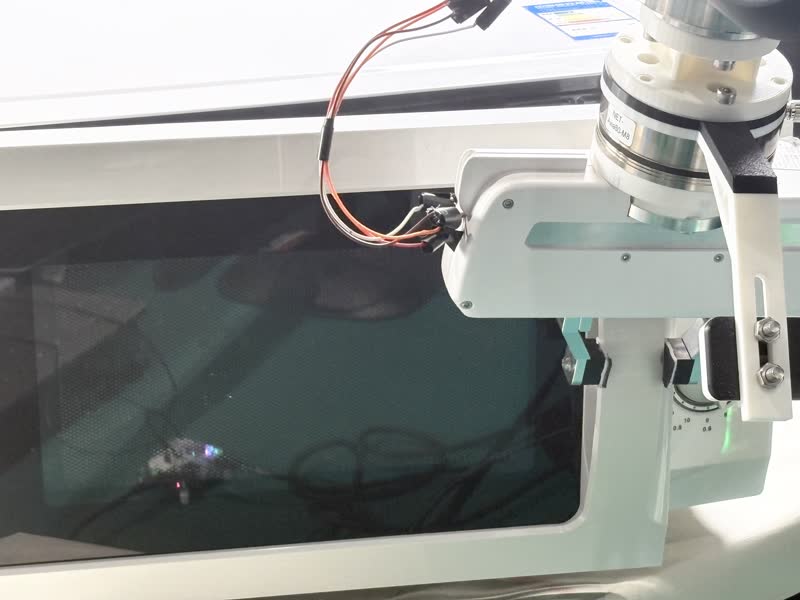}\\[3pt]
            \raggedright
            \textbf{Task:} Microwave\\  
            \textbf{Instruction:} Open the door of the microwave.\\  
            \textbf{Description:} Grip the microwave door handle and pull the door open, allowing it to rotate about its hinge.
        \end{minipage}
        &
        \begin{minipage}[t]{4.45cm}
            \centering
            \includegraphics[width=2.0cm]{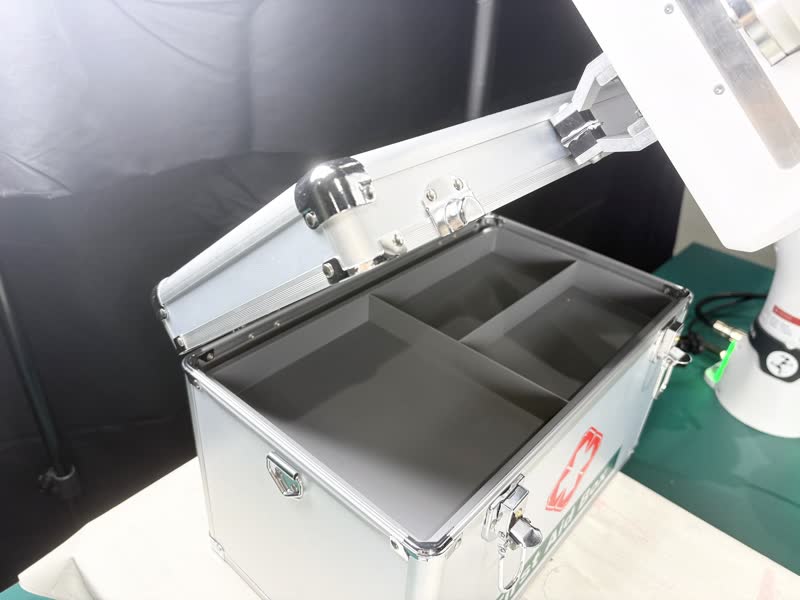}\\[3pt]
            \raggedright
            \textbf{Task:} Open-Box-1\\  
            \textbf{Instruction:} Push the lid from bottom left to the top right.\\  
            \textbf{Description:} Contact the side of the first-aid kit lid with the gripper, then push it from bottom-left toward top-right to open it about its hinge.
        \end{minipage}
        &
        \begin{minipage}[t]{4.45cm}
            \centering
            \includegraphics[width=2.0cm]{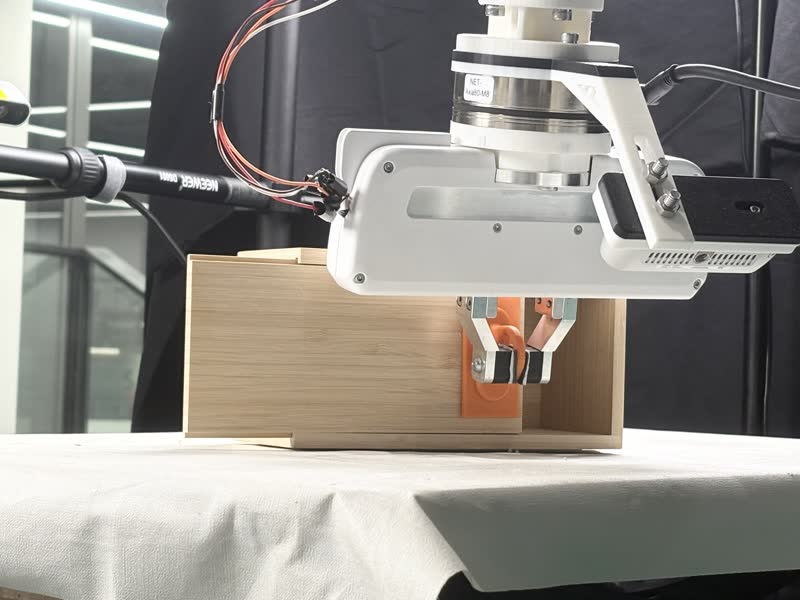}\\[3pt]
            \raggedright
            \textbf{Task:} Open-Box-2\\  
            \textbf{Instruction:} Slide the box lid leftward to open it.\\  
            \textbf{Description:} Grip the box lid handle, then slide the lid leftward in parallel to open it.
        \end{minipage}
        &
        \begin{minipage}[t]{4.45cm}
            \centering
            \includegraphics[width=2.0cm]{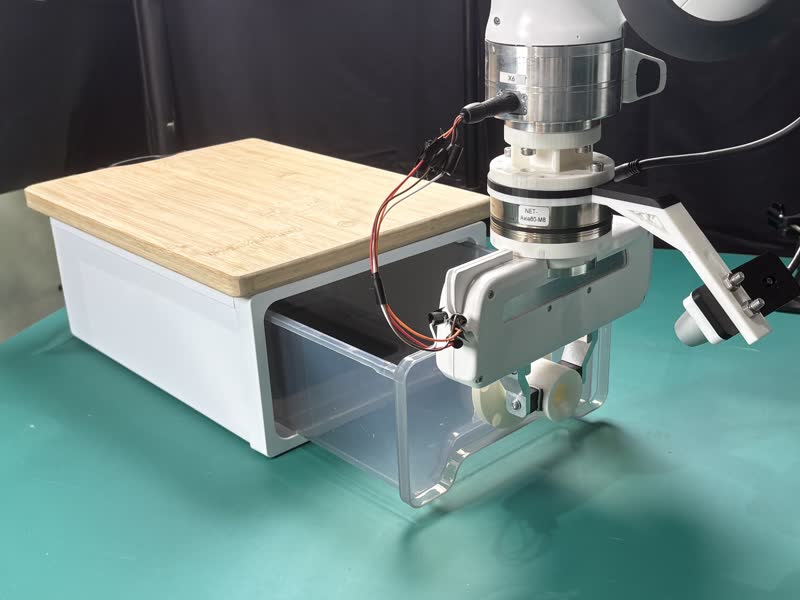}\\[3pt]
            \raggedright
            \textbf{Task:} Pull-Drawer\\  
            \textbf{Instruction:} Pull the drawer.\\  
            \textbf{Description:} Grip the drawer handle, then pull the drawer open in a parallel motion along its rails.
        \end{minipage}
        \\

        \bottomrule
    \end{tabular}
\end{table*}

\subsection{Model}
\subsubsection{Architecture}
\paragraph{Action and Force Encoding}
Future actions are embedded by a two-layer MLP with sinusoidal positional encodings. A temporal convolutional network (TCN) encodes the wrench history, augmented with first-order differences and wrench magnitudes, into temporally resolved contact features. The resulting features provide a sequence of dynamic tokens, while their temporal mean and standard deviation are mapped to a global contact-state token. These complementary representations supply the temporal feedback and global modulation used by the action–force context builder.

\paragraph{Predictive Readout and Anomaly Head} A temporal transformer followed by an MLP decodes interaction tokens into future wrench predictions. The anomaly head encodes predicted visual latents with a compact CNN and projects interaction tokens and predicted forces through separate MLPs. Their temporally aligned features are fused into a 128-dimensional representation at each future step and processed by a two-layer, four-head temporal transformer to model dependencies across the prediction horizon. The window classifier aggregates the top 20\% of temporal features ranked by the auxiliary step head scores, emphasizing localized anomalous evidence. The prevalence head instead aggregates all temporal features to estimate the extent of anomalies over the prediction horizon.

\begin{table*}[t]
\centering
\caption{Training hyperparameters for MAIA.}
\label{tab:training_hyperparams}
\begin{tabular}{lccccl}
\toprule
% Stage & Steps & Batch (per-GPU) & LR & Warmup & Loss weights \\
\multicolumn{1}{c}{\textbf{Stage}} &
\multicolumn{1}{c}{\textbf{Steps}} &
\multicolumn{1}{c}{\textbf{Batch (per-GPU)}} &
\multicolumn{1}{c}{\textbf{LR}} &
\multicolumn{1}{c}{\textbf{Warmup}} &
\multicolumn{1}{c}{\textbf{Loss weights}} \\

\midrule
Stage 1: VAF-WM &
30k & 8 & $3{\times}10^{-5}$ & 1k &

$\lambda_{\mathrm{v}}=1.0$, $\lambda_{\mathrm{f}}=1.0$, $\lambda_{\Delta \mathrm{f}}=0.2$, $\lambda_{\mathrm{smooth}}=0.1$, $\lambda_{\mathrm{sem}}=0.05$ \\
\midrule
Stage 2: Anomaly Head &
5k & 32 & $5{\times}10^{-5}$ & 500 &
$\lambda_{\mathrm{y}}=1.0$, $\lambda_{\mathrm{p}}=0.5$, $\lambda_{\mathrm{a}}=0.5$ \\
\midrule
Stage 3: Novel-Task Adaptation &
2k & 8 & $1{\times}10^{-5}$ & 500 &
$\lambda_{\mathrm{y}}=1.0$, $\lambda_{\mathrm{p}}=0.5$, $\lambda_{\mathrm{a}}=0.5$, $\lambda_{\mathrm{force}}^{\mathrm{nom}}=0.1$ \\
\bottomrule
\end{tabular}
% \vspace{-8pt}
\end{table*}

\begin{table}[t]
    \centering
    \begin{threeparttable}
    \caption{Task-level anomaly anticipation performance across all 18 tasks.}
    \label{tab:task_level_anomaly}
    % \setlength{\tabcolsep}{6pt}
    % \vspace{-2pt}
    \begin{tabular}{lcccc}
        \toprule
        \textbf{Category / Task} &
        \textbf{AUPRC$\uparrow$} &
        \textbf{bACC$\uparrow$} &
        \textbf{AR$\uparrow$} &
        \textbf{Med. ALT$\uparrow$} \\
        \midrule

        \multicolumn{5}{l}{\textit{Insertion}} \\
        \quad Insert-Plug-2    & 0.949 & 0.887 & 0.581 & 27.0 \\
        \quad Insert-Plug-3    & 0.963 & 0.912 & 0.567 & 39.5 \\
        \quad Insert-Type-C     & 0.973 & 0.804 & 0.412 & 20.0 \\
        \quad Insert-NI$^*$    & 0.962 & 0.832 & 0.470 & 25.0 \\
        \midrule

        \multicolumn{5}{l}{\textit{Wiping}} \\
        \quad Wipe-Board       & 0.716 & 0.797 & 0.600 & 30.0 \\
        \quad Wipe-Bowl        & 0.562 & 0.689 & 0.428 & 32.0 \\
        \quad Wipe-Vase$^*$    & 0.556 & 0.683 & 0.434 & 22.0 \\
        \midrule

        \multicolumn{5}{l}{\textit{Pressing}} \\
        \quad Press-Seal       & 0.545 & 0.686 & 0.278 & 19.5 \\
        \quad Press-Switch-1   & 0.706 & 0.645 & 0.667 & 17.0 \\
        \quad Press-Switch-2   & 0.786 & 0.776 & 0.346 & 38.0 \\
        \quad Press-Bottle$^*$ & 0.854 & 0.662 & 0.406 & 19.5 \\
        \midrule

        \multicolumn{5}{l}{\textit{Path-Constrained Manipulation}} \\
        \quad Push-I           & 0.705 & 0.799 & 0.432 & 22.5 \\
        \quad Zipper           & 0.586 & 0.727 & 0.400 & 29.5 \\
        \quad Push-C$^*$       & 0.718 & 0.743 & 0.351 & 30.0 \\
        \midrule

        \multicolumn{5}{l}{\textit{Articulated-Object Manipulation}} \\
        \quad Microwave        & 0.747 & 0.704 & 0.267 & 38.0 \\
        \quad Open-Box-1       & 0.951 & 0.901 & 0.643 & 19.0 \\
        \quad Open-Box-2       & 0.596 & 0.640 & 0.417 & 31.0 \\
        \quad Pull-Drawer$^*$  & 0.550 & 0.631 & 0.381 & 29.0 \\
        \bottomrule

        % \textbf{Seen Average}
        %     & \textbf{0.738} & \textbf{0.765}
        %     & \textbf{0.464} & \textbf{28.0} \\
        % \textbf{Novel Average}
        %     & \textbf{0.728} & \textbf{0.710}
        %     & \textbf{0.408} & \textbf{25.1} \\
        % \bottomrule
    \end{tabular}
    \begin{tablenotes}
    \footnotesize
    \item $^*$ denotes novel tasks.
    \end{tablenotes}
    \end{threeparttable}
    \vspace{-6pt}
\end{table}

\paragraph{Language-Guided Interaction Regularization}
We encourage interaction representations to preserve the semantic relationships between task descriptions through an auxiliary relational loss:
\begin{equation}
    L_{sem}= \frac{1}{B(B-1)}\sum_{i \neq j}\left(\mathbf{u}_i^\top \mathbf{u}_j - \operatorname{sg}\left[\mathbf{v}_i^\top \mathbf{v}_j\right] \right)^2.
\end{equation}
Here, $\mathbf{u}$ and $\mathbf{v}$ denote normalized projections of pooled interaction and task-language features, respectively, and $\operatorname{sg}$ denotes stop-gradient. The loss matches pairwise similarities rather than individual features, preserving task-level semantic structure without enforcing identical representations across modalities. It is applied during nominal world-model training to samples with retained language conditioning, with weight 0.05; \(B\) denotes the number of eligible samples in the batch.

\subsubsection{Training}
We use the same temporal sampling protocol described in the main paper and report here the remaining implementation details needed for reproducibility. All stages are trained with AdamW, bfloat16 mixed precision, gradient clipping with maximum norm 1.0, and a constant learning-rate schedule with linear warmup. Unless otherwise specified, we use AdamW betas $(0.9,0.95)$, weight decay $10^{-5}$, and random seed 42. Action and force/torque signals are normalized using statistics computed from the training split. All reported runs are trained on 4 NVIDIA A100 GPUs. The stage-specific optimization settings are summarized in Table~\ref{tab:training_hyperparams}.

During VAF-WM training, the VAE and text encoder are frozen, while the video transformer and action-force pathway are jointly optimized. Task-language conditioning is dropped with probability 0.06.
For anomaly-head training, VAF-WM is frozen and rollout features are generated with a single inference step, matching deployment. The anomaly head is trained with a positive class weight of 5.0 to address class imbalance.
During novel-task adaptation, the video backbone remains frozen, while the action-force pathway and anomaly head are jointly fine-tuned. Nominal force regularization is applied only to nominal adaptation windows.

% During VAF-WM training, the VAE and text encoder are kept frozen, while the video transformer and the action-force pathway are optimized jointly. Task-language conditioning is dropped with probability 0.06. The world model is trained on nominal windows only, so that the learned dynamics represent expected interaction behavior rather than failure-specific patterns.

% For anomaly-head training, VAF-WM is frozen and rollout features are generated with a single inference step, matching the deployment setting. The anomaly head is then optimized on both nominal and anomalous windows, using a positive class weight of 5.0 to address class imbalance.

% For novel-task adaptation, we initialize from the trained VAF-WM and anomaly head. The video backbone remains frozen, while the action-force pathway and anomaly head are fine-tuned. The nominal force regularization is evaluated only on nominal adaptation windows, which helps preserve the learned nominal interaction dynamics while adapting the anomaly predictor to task-specific failure patterns.

\subsection{Supplementary Experimental Results}
\begin{figure}[t]
    \centering
    % \vspace{-4pt}
    \includegraphics[width=\linewidth]{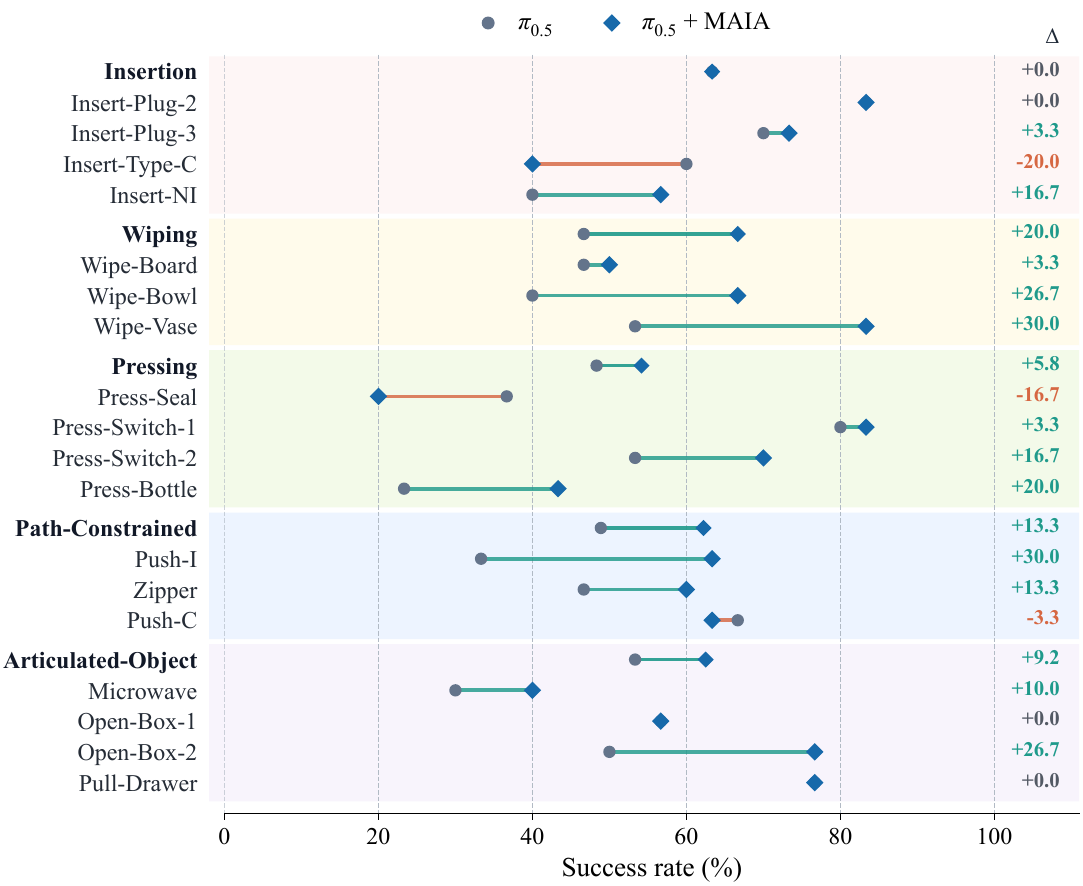}
    \caption{Comparison of the original $\pi_{0.5}$ policy and $\pi_{0.5}$ with MAIA across 18 contact-rich manipulation tasks. The rightmost column shows the success-rate difference in percentage points.}
    \label{fig:task_level_success}
    \vspace{-6pt}
\end{figure}

Table~\ref{tab:task_level_anomaly} reports task-level anomaly anticipation performance across all 18 tasks, while Fig.~\ref{fig:task_level_success} presents the real-robot execution success rates with and without MAIA. Together, these results provide a finer-grained breakdown of the aggregate evaluations reported in the main paper.
% \subsubsection{Task-Level Anomaly Anticipation Performance}
% Table~\ref{tab:task_level_anomaly} reports the task-level anomaly anticipation results across all 18 tasks in CRIA. We report AUPRC, balanced accuracy (bACC), anticipation rate (AR), and median anticipation lead time (ALT). These results complement the aggregate performance reported in the main paper by providing a finer-grained view of performance variations across individual tasks and interaction categories.

% \subsubsection{Task-Level Real-Robot Policy Integration}
% Fig.~\ref{fig:task_level_success} presents the task-level success rates of the original $\pi_{0.5}$ policy and its integration with MAIA across all 18 tasks. 
% This breakdown complements the category-level results in the main paper and illustrates how anomaly-aware action selection affects execution success across individual tasks.

